\documentclass[lettersize,journal]{IEEEtran}
\usepackage{amsmath,amsfonts}
\usepackage{array}
\usepackage[caption=false,font=normalsize,labelfont=sf,textfont=sf]{subfig}
\usepackage{textcomp}
\usepackage{stfloats}
\usepackage{url}
\usepackage{verbatim}
\usepackage{graphicx}
\usepackage{cite}
\usepackage{xcolor}
\usepackage[ruled, linesnumbered, noend]{algorithm2e}

\usepackage{float}

\usepackage{pifont}

\usepackage[OT1]{fontenc} 
\usepackage{balance}
\begin{document}

\title{BiAIT*: Symmetrical Bidirectional Optimal Path Planning with Adaptive Heuristic}

\author{Chenming Li,
        Han Ma,
        Peng Xu,
        Jiankun Wang,~\IEEEmembership{Member,~IEEE},
        Max Q.-H. Meng,~\IEEEmembership{Fellow,~IEEE}
        \thanks{This project is supported by National Key R\&D program of China (No. 2019YFB1312400) and Hong Kong RGC GRF (\# 14211420) awarded to Max Q.-H. Meng. \textit{(Corresponding authors: Jiankun Wang, Max Q.-H. Meng.)}}
        \thanks{Chenming Li, Han Ma, and Peng Xu are with the Department of Electronic Engineering, The Chinese University of Hong Kong, Shatin, N.T., Hong Kong SAR, China (e-mail: licmjy@link.cuhk.edu.hk; hanma@link.cuhk.edu.hk; peterxu@link.cuhk.edu.hk).}
        \thanks{Jiankun Wang is with Shenzhen Key Laboratory of Robotics Perception and Intelligence, and the Department of Electronic and Electrical Engineering, Southern University of Science and Technology, Shenzhen 518055, China (e-mail: wangjk@sustech.edu.cn).}
        \thanks{Max Q.-H. Meng is with Shenzhen Key Laboratory of Robotics Perception and Intelligence, and the Department of Electronic and Electrical Engineering, Southern University of Science and Technology, Shenzhen 518055, China, on leave from the Department of Electronic Engineering, The Chinese University of Hong Kong, Hong Kong, and also with the Shenzhen Research Institute of The Chinese University of Hong Kong, Shenzhen 518057, China, max.meng@ieee.org.} 
        
        }
        
\markboth{Journal of \LaTeX\ Class Files,~Vol.~14, No.~8, August~2021}%
{Shell \MakeLowercase{\textit{et al.}}: A Sample Article Using IEEEtran.cls for IEEE Journals}


\maketitle

\begin{abstract}
    Adaptively Informed Trees (AIT*) is an algorithm that uses the problem-specific heuristic to avoid unnecessary searches, which significantly improves its performance, especially when collision checking is expensive. 
    However, the heuristic estimation in AIT* consumes lots of computational resources, and its asymmetric bidirectional searching strategy cannot fully exploit the potential of the bidirectional method.
    In this article, we propose an extension of AIT* called BiAIT*. 
    Unlike AIT*, BiAIT* uses symmetrical bidirectional search for both the heuristic and space searching.
    The proposed method allows BiAIT* to find the initial solution faster than AIT*, and update the heuristic with less computation when a collision occurs. 
    We evaluated the performance of BiAIT* through simulations and experiments, and the results show that BiAIT* can find the solution faster than state-of-the-art methods. 
    We also analyze the reasons for the different performances between BiAIT* and AIT*. 
    Furthermore, we discuss two simple but effective modifications to fully exploit the potential of the adaptively heuristic method.
\end{abstract}

\def\abstractname{Note to Practitioners}
\begin{abstract}
    This work is inspired by the adaptively heuristic method and the symmetrical bidirectional searching method. 
    The article introduces a novel algorithm that uses the symmetrical bidirectional method to calculate the adaptive heuristic and efficiently search the state space. 
    The problem-specific heuristic in BiAIT* is derived from a lazy-forward tree and a lazy-reverse tree, which are constructed without collision checking. 
    The lazy-forward and lazy-reverse trees are enabled to meet in the middle, thus generating the effective and accurate heuristic.
    In BiAIT*, the lazy-forward and lazy-reverse trees share heuristic information and jointly guide the growth of the forward and reverse trees, which conduct collision checking and guarantee the feasibility of their edges. 
    Compared with state-of-the-art methods, BiAIT* finds the initial heuristic and updates the heuristic more quickly.
    The proposed algorithm can be applied to industrial robots, medical robots, or service robots to achieve efficient path planning. 
    The implementation of BiAIT* is available at {https://github.com/Licmjy-CU/BiAITstar}.
\end{abstract}


\section{Introduction}

\IEEEPARstart{P}{lanning} a feasible trajectory is one of the key problems in robotics.
This task can be simplified as finding a collision-free geometric path for the robot to follow. 
Nevertheless, the geometric path planning problem is still computationally expensive, especially in high-dimensional space.
Usually, the most time-consuming procedure in path planning is collision checking.
Various methods have been proposed to solve the robot path planning problem.


The planning problem in discrete low-dimensional space can be solved using search-based methods such as Dijkstra \cite{dijkstra1959note}, A* \cite{hart1968formal}, and Lifelong Planning A* (LPA*) \cite{koenig2004lifelong}. 
However, the computational complexity of search-based methods grows exponentially as the dimension increases.
Sampling-based methods like Rapidly-exploring Random Tree (RRT) \cite{lavalle1998rapidly} and its advanced version, RRT* \cite{karaman2010incremental}, can efficiently search the state space. 
RRT and RRT* use random sampling to avoid explicit construction of the configuration space. 
The typical sampling-based planner consists of a sampling stage and a searching stage. 
In the sampling stage, the planner samples the entire state space to provide a topological abstraction of the space. 
Then, in the searching stage, the planner searches on the topological graph. 
RRT and RRT* maintain a tree to search the space and alternately perform the sampling and searching stages until they find a satisfactory path or the computation time reaches the threshold and a failure is reported.

\begin{figure}[htbp]
    \flushright
    \includegraphics[width=0.95\linewidth]{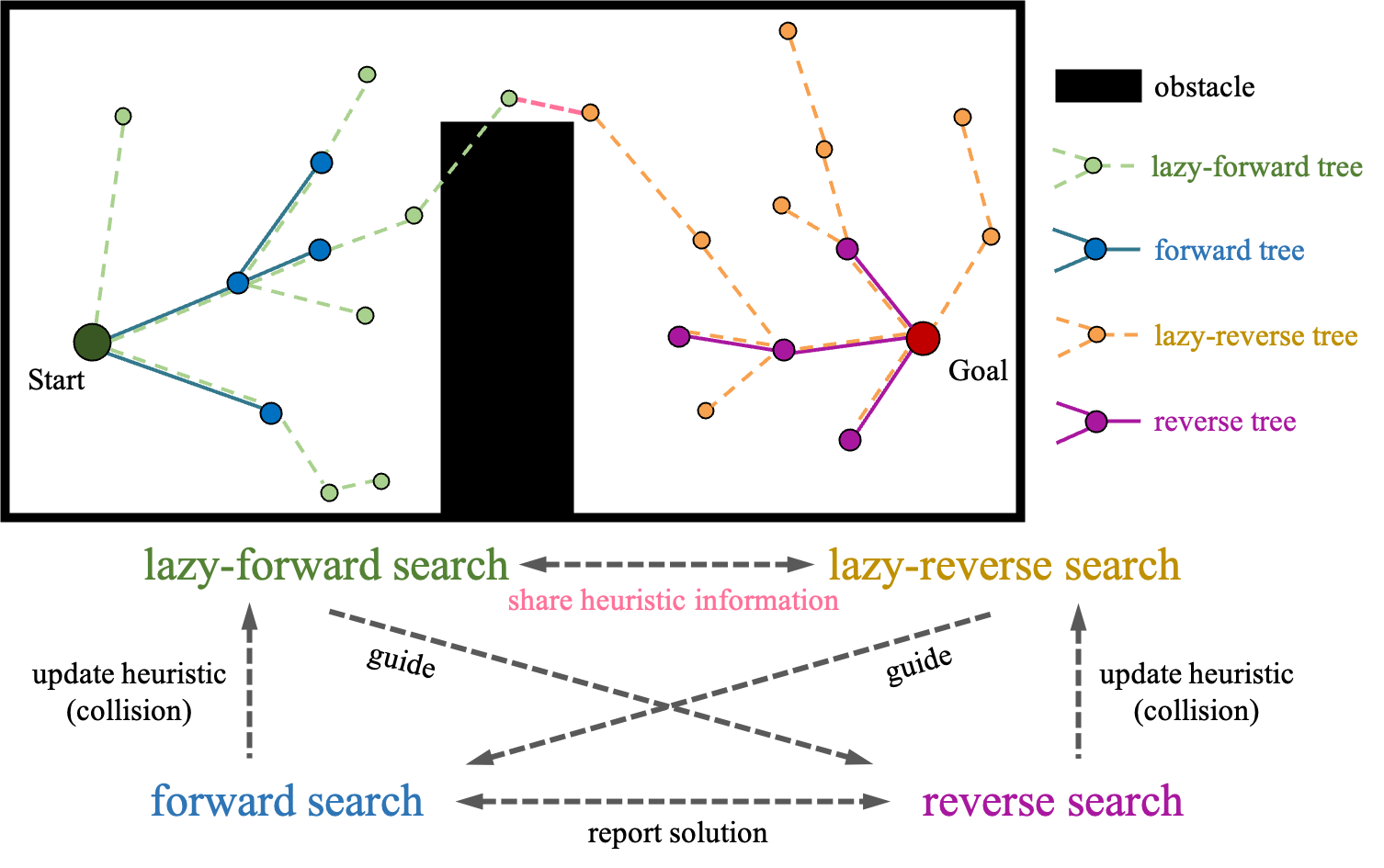}
    
    \caption{ The components of BiAIT*. 
    The lazy-forward and lazy-reverse trees provide the problem-specific heuristic to guide the forward and reverse searches.
    The lazy trees are constructed without collision checking, and they share heuristic information when they meet in the middle.
    The forward and reverse trees search the space according to the heuristic and guarantee that their edges are feasible with collision checking.
    If forward and reverse searches find the edge in lazy trees collides with the obstacle, they will inform the lazy trees to update the heuristic.
    When forward and reverse trees meet, BiAIT* will report a new solution.
    }
    \label{SchematicFigure}
\end{figure}

\begin{figure*}[!htb]
    \centering
    \subfloat[]{\includegraphics[width=0.19\linewidth]{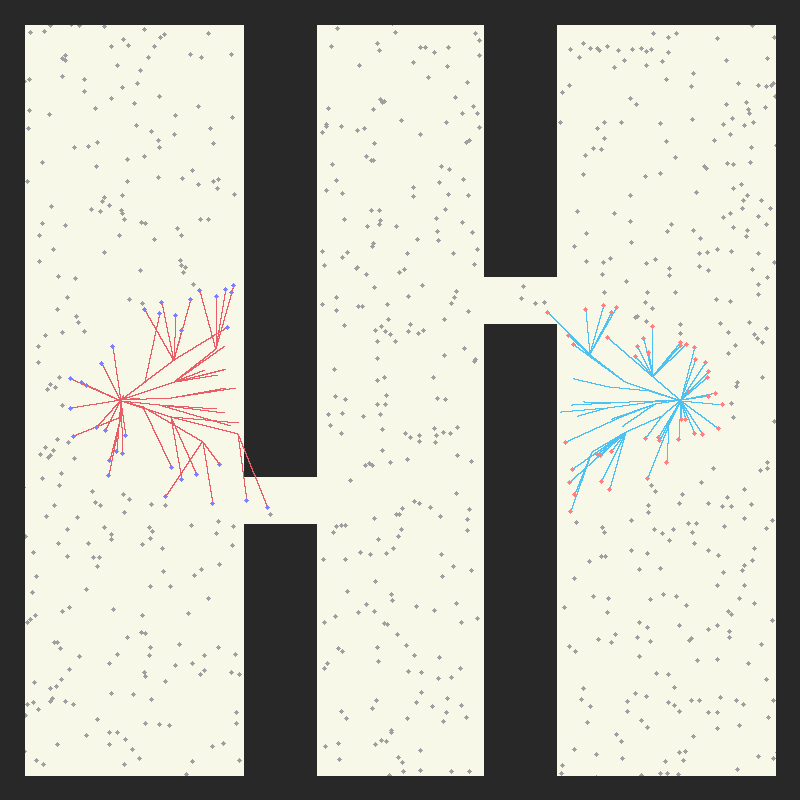}%
    \label{D1}}
    \hfil
    \subfloat[]{\includegraphics[width=0.19\linewidth]{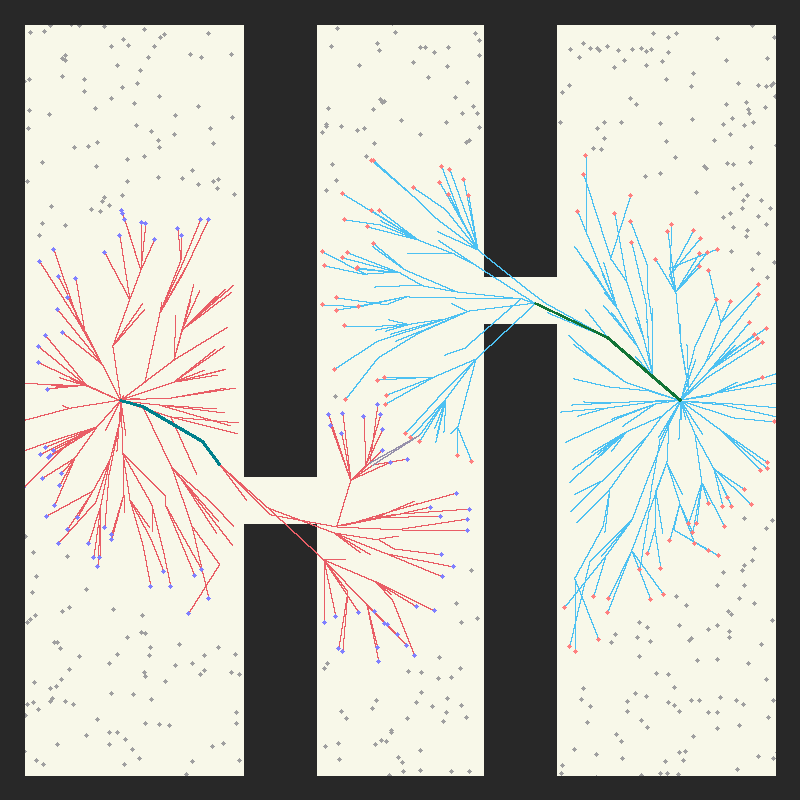}%
    \label{D2}}
    \hfil
    \subfloat[]{\includegraphics[width=0.19\linewidth]{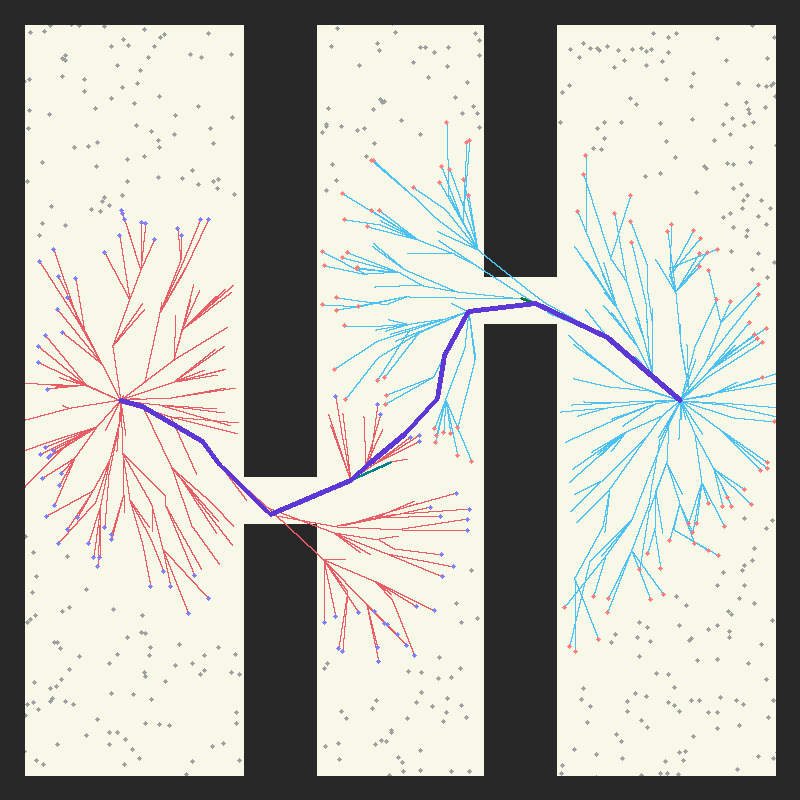}%
    \label{D3}}
    \hfil
    \subfloat[]{\includegraphics[width=0.19\linewidth]{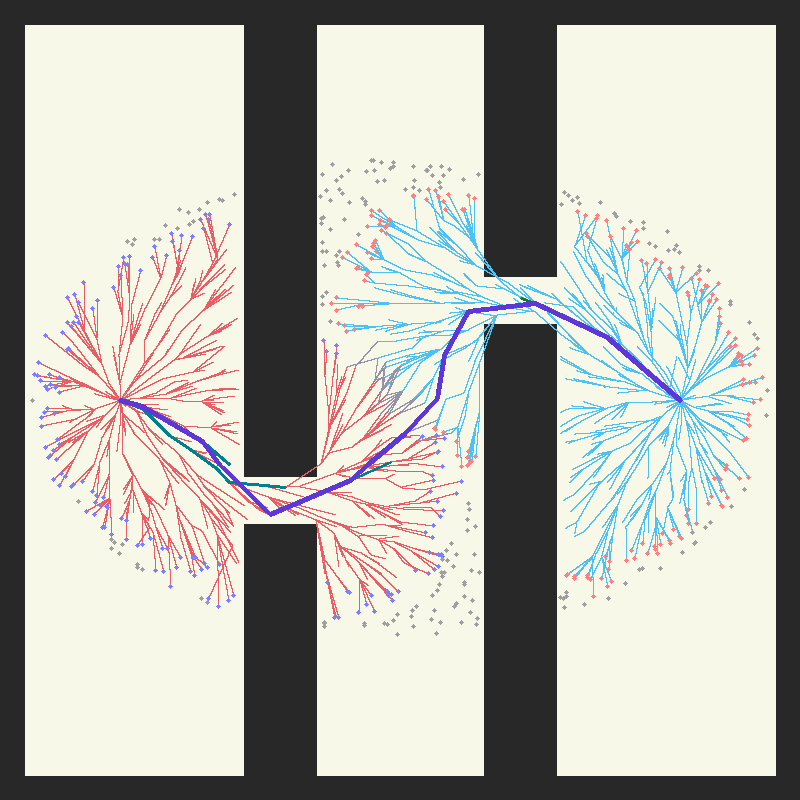}%
    \label{D4}}
    \hfil
    \subfloat[]{\includegraphics[width=0.19\linewidth]{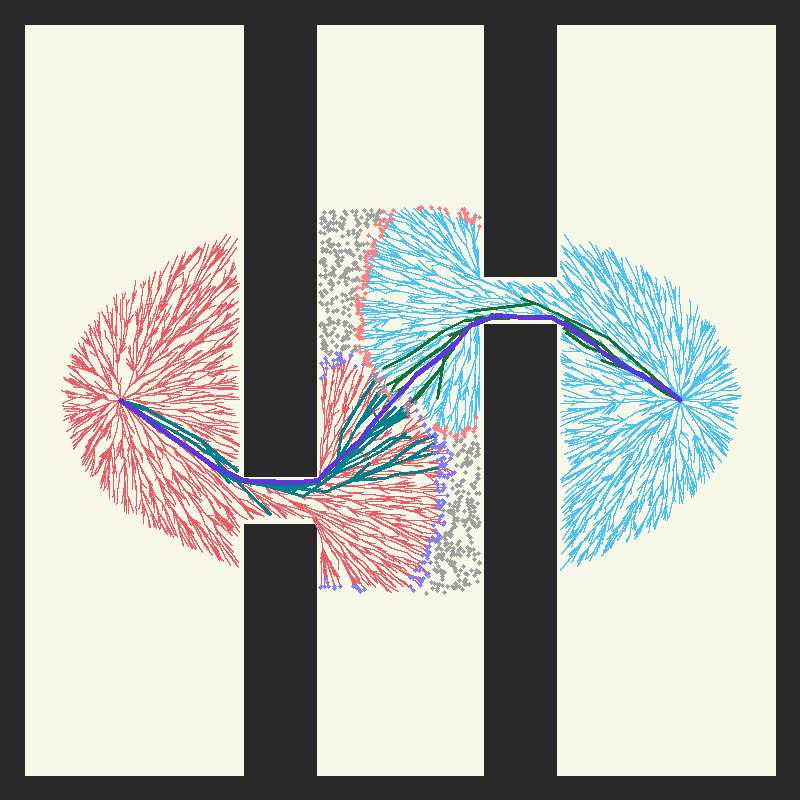}%
    \label{D5}}

    
    \caption{ BiAIT* solves a planning problem in a simple 2D environment, where the black blocks, red lines, blue lines, green lines, violet line denote the obstacles, lazy-forward tree, lazy-reverse tree, forward and reverse trees, and the current solution path, respectively. 
    The grey points are the idle samples. 
    The blue and red points are the wavefronts of the lazy-forward and lazy-reverse searches.
    Fig. \ref{D1}: BiAIT* takes a batch of samples and constructs the lazy-forward and lazy-reverse trees; Fig. \ref{D2}: the lazy-forward and lazy-reverse trees meet in the middle and share the heuristic information, and the forward and reverse trees start searching the space according to the heuristic; Fig. \ref{D3}: the initial feasible solution; Fig. \ref{D4}: BiAIT* prunes the samples and branches outside the Informed space and starts a new iteration; Fig. \ref{D5}: BiAIT* keeps updating the solution until it meets the terminate condition.
    }
    \label{PlanningProcedureFigure}
\end{figure*}


RRT and RRT* sample uniformly and search omnidirectionally, which often leads to the exploration of lots of redundant space. 
Batch Informed Trees (BIT*) \cite{gammell2015batch, gammell2020batch} separates the sampling and searching stages into different phases instead of performing them alternately. 
BIT* constructs the tree using ordered samples stored in a priority queue. 
Adaptively Informed Trees (AIT*) \cite{strub2020adaptively, strub2021ait} is developed based on BIT*, which searches the space with adaptive heuristics. 
Unlike conventional bidirectional search, AIT* maintains a lazy-reverse tree and a forward tree. 
AIT* takes a batch of samples in the sampling stage and constructs a lazy-reverse tree. 
The lazy-reverse tree, which does not guarantee that its edges are feasible, provides the problem-specific heuristic for the state space topological abstraction. 
This problem-specific heuristic can provide better guidance than the Euclidean metric, and the forward tree rooted at the start only needs to extend a small portion of the state space. 
However, AIT* consumes significant computational resources for problem-specific heuristic estimation. 
AIT* is an asymmetric planning method, and its lazy-reverse tree has to extend until it meets the start state to generate the initial heuristic. 
When a collision occurs, the whole lazy-reverse tree branch rooted at the collided point will be invalidated, and AIT* needs to reconstruct the invalidated components.

We propose to improve AIT* with the symmetrical bidirectional method, namely Bidirectional Adaptively Informed Trees (BiAIT*).
BiAIT* reconstructs the forward tree and lazy-reverse tree in AIT* into four parts: a forward tree, a lazy-forward tree, a reverse tree, and a lazy-reverse tree.
BiAIT* builds its forward and reverse trees with feasible edges, and lazily builds the lazy-forward and lazy-reverse trees without collision checking.
Fig. \ref{SchematicFigure} shows the functions of these four parts in BiAIT*.
Our method constructs the lazy-forward tree and lazy-reverse tree in the beginning of each batch of samples.
The lazy-forward and lazy-reverse trees meet in the middle of resolution-optimal heuristic path under the current sample set.
When lazy trees meet, they inform each other and propagate the problem-specific heuristic information towards their predecessors.
The forward and reverse trees extend according to the problem-specific heuristic. 
During the forward and reverse search procedures, BiAIT* extracts the best candidate edge from the edge queues and checks whether it is collision-free.
If the candidate edge collides with the obstacles, BiAIT* informs the lazy trees and updates the heuristic. 
BiAIT* terminates the searches when they cannot improve the current trees or the solution, and samples a new batch.
With the symmetrical bidirectional adaptive heuristic, BiAIT* can acquire the initial heuristic by extending fewer vertices in its lazy search than AIT*.
In addition, BiAIT* reduces the workload of updating the heuristic when collisions occur.
Fig. \ref{PlanningProcedureFigure} shows the schematic and planning procedure of our method, and we will introduce the details in the following sections.

The heuristic in BiAIT* can be viewed from two perspectives.
The first perspective is the adaptive heuristic derived from the lazy-forward search and the lazy-reverse search. 
The second perspective is the symmetric search strategy, which simultaneously grows two normal trees and two lazy trees from the start and the goal. 
In this way, the effect of the adaptive heuristic is further improved. 
Searching from the start and the goal helps the planner solve the narrow passage problem more easily than AIT*.
The process of lazy-forward search and lazy-reverse search is more computationally efficient, as the expanded region is smaller and the heuristic updating process takes less workload. 

Our contributions can be summarized as follows: 1) We proposes BiAIT*, which extends AIT* from an asymmetric bidirectional mode to the symmetrical bidirectional mode. 2) We comprehensively evaluate the performance of BiAIT* in different environments and demonstrate that the symmetrical bidirectional adaptively informed method can improve the planning speed. 3) We discuss two simple but effective modifications of AIT* and BiAIT*, which improve their performance with almost no additional consumption.

We organize this article as follows. 
In Section II, we review related literatures. 
Section III defines the problem formally and provides a brief review of AIT*.
Section IV illustrates the details of BiAIT*.
Section V presents the results of the simulations.
In section VI, we present three real-world experiments.
We discuss the effectiveness and limitation of BiAIT* in Section VII.
We propose two simple extensions of AIT* and BiAIT* in Section VIII.
Finally, we draw the conclusion in Section IX.

\section{Related Work}

\subsection{Planning Algorithms}

The sampling-based method avoids explicitly discretizing the state space, making it possible to solve the continuous planning problem without space discretization.
Methods like RRT \cite{lavalle1998rapidly} can guarantee the probabilistic completeness, which means the probabilistic of finding a solution goes to one when the number of samples goes to infinity \cite{wang2021survey}.
Nevertheless, RRT can not guarantee the optimal property, and its solution will not converge to the optimal.
It has been proved that RRT has zero chance to find the optimal solution, even with an infinite number of samples \cite{karaman2011sampling}.
RRT* \cite{karaman2010incremental} adds two processes, which are finding the best parent and rewiring.
RRT* can acquire the asymptotic optimality, which means the chance of finding the optimal solution goes to one with infinite samples.
RRT$^{\#}$ \cite{arslan2013use} improves RRT* with a replanning procedure.
When the path cost is defined with the Euclidean metric, Informed RRT* \cite{gammell2014informed} uses a direct sampling method to sample in the informed space after a feasible path is found and proves that sampling outside the informed space can not promote the current solution.
The informed space is a $d$-dimensional prolate hyper-spheroid.
Moreover, Gammell \emph{et al.} \cite{gammell2018informed} propose pruning the branch outside the informed space.
The Informed RRT* achieves a better convergence rate than RRT*.
The artificial neural network can guide the sampling process and improve sampling efficiency.
Neural RRT* \cite{wang2020neural} learns from the A* algorithm with a CNN model and predicts the distribution of the optimal path, which serves as a heuristic for sampling.
However, calculating an accurate and computationally inexpensive heuristic is still challenging. 
Selecting an appropriate heuristic for a specific planning problem is also problematic.

\subsection{Batch Sampling Methods}

The sampling-based planning method has two stages: sampling and searching stages.
Methods like RRT and RRT* alternately perform the sampling stage and the searching stage.
However, the random order of samples is inefficient and results in unnecessary computation.
Inspired by the Fast Marching method, Janson \emph{et al.} proposes Fast Marching Tree (FMT*) \cite{janson2015fast}.
FMT* takes a set of samples in the sampling stage and performs forward dynamic programming recursion in the searching stage.
The searching tree of FMT* grows steadily outward and never checks the vertices in the closed set.
FMT* achieves significant improvement compared with RRT*.
However, FMT* is not an anytime algorithm since it runs on a fixed topological abstraction and never updates its samples.
Batch Informed Trees (BIT*) \cite{gammell2015batch, gammell2020batch} takes a batch of samples in the informed space in each iteration, building an increasing dense approximation of the state space.
Besides, BIT* uses priority queues to order the samples, and the priority queues are sorted concerning the potential to improve the current solution.
Regionally accelerated BIT* \cite{choudhury2016regionally} extends BIT* with a local optimizer and improves the performance in difficult-to-sample planning problems, such as the problem of the narrow passage.
AIT* \cite{strub2020adaptively, strub2021ait} is developed based on BIT*, which utilizes a lazy-reverse tree to calculate the adaptively cost-to-go heuristic for the samples. 
AIT* addresses that the priority queue in BIT* is sorted without considering the specific planning problem.
In each iteration, AIT* first takes a batch of samples and then builds the lazy-reverse tree to get the adaptively heuristic.
The priority edge queue in AIT* is sorted with the adaptive heuristic.
The related branch in the lazy-reverse tree will be updated when the collision happens.

\subsection{Bidirectional Methods}

Typical bidirectional planning methods utilize a forward-searching tree and a reverse-searching tree to search the space and extend these two trees alternately.
RRT-Connect \cite{kuffner2000rrt} first extends RRT with the bidirectional searching strategy, and it uses a greedy connect function to find the solution quickly.
RRT-Connect can achieve better performance than RRT in various path planning problems.
RRT*-Connect \cite{klemm2015rrt} combines RRT* and RRT-Connect, and successfully guarantees both the asymptotically optimality like RRT* and the capability of quickly finding solutions like RRT-Connect.
Informed RRT*-Connect \cite{mashayekhi2020informed} uses RRT*-Connect to find the feasible solution and constraints the samples in the informed space.
Wang \emph{et al.} propose a novel method \cite{wang2021efficient} to connect two trees in bidirectional motion planning, which could avoid solving the two-point boundary value problem in the connection process.
Ma \emph{et al.} propose Bi-Risk-RRT \cite{ma2022bi}, which extends Risk-RRT \cite{rios2011understanding} with a bidirectional search method.
Bi-Risk-RRT proves the bidirectional method is efficient when considering kinodynamic constraints.
Bidirectional FMT* \cite{starek2014bidirectional} extends the FMT* to bidirectional search, which performs bidirectional dynamic programming on a set of samples.
AIT* \cite{strub2021ait} uses the asymmetric bidirectional searching method to solve the path planning problem. 
The reverse tree of AIT* is different from the one in typical bidirectional path planning methods like RRT-Connect and RRT*-Connect. 
In AIT*, only the forward tree is responsible for searching the space, and the reverse tree provides an adaptively heuristic for the forward tree.
Viewed separately, the forward and lazy-reverse searches in AIT* are two unidirectional procedures instead of a typical bidirectional procedure.
The asymmetric bidirectional method in AIT* can not fully exploit the potential of the bidirectional method.

Bidirectional heuristic search improves the efficiency of bidirectional methods by avoiding to search irrelevant regions with heuristic. 
However, conventional bidirectional heuristic search typically generates suboptimal solutions since its forward and reverse searches tend to explore along different paths that fail to meet in the middle.
The Meet-in-the-Middle (MM) \cite{holte2016bidirectional} algorithm proposes constraining the forward and reverse searches within the bidirectional brute-force searched regions, guaranteeing an optimal solution. 
However, MM and its successors \cite{sturtevant2018brief}, \cite{holte2017mm} are designed for low-dimensional discrete space searching problems and are not efficient path planning methods in high-dimensional continuous space.

\section{Preliminaries}

\subsection{Problem Formulation}
Sampling-based methods are often evaluated in two aspects: probabilistic completeness and almost-surely asymptotical optimality.
Most sampling-based planning methods like RRT \cite{lavalle1998rapidly} and RRT* \cite{karaman2010incremental} can guarantee probabilistic completeness, which means the planner will find a feasible solution eventually if the solution exists.

Let the $d$-dimensional space be $\Re^d$.
$\mathcal{X}$ denotes the state space, which is a subset of $\Re^d$.
$\mathcal{X}_{obs}$ represents the obstacle space. 
We can define $\mathcal{X}_{free} = \mathcal{X} \backslash \mathcal{X}_{obs}$ as the free space. 
Let $\pi: [0, 1] \to \mathcal{X}_{free}$ be a feasible path lying in $\mathcal{X}_{free}$, and $\Pi$ be the collection of all feasible paths.
Define the cost function as $c(\pi) \to [0, \infty)$.
Then, the optimal path planning problem can be summarized as:
\begin{equation}
       \pi^{*} = \operatorname*{\arg \min}_{\pi \in \Pi}\{c(\pi) | \pi(0) = x_{s}, \pi(1) \in \mathcal{X}_{g}\},
\end{equation}
where the $x_s$ and $\mathcal{X}_{g}$ represent the start state and the goal region, respectively.
The almost-surely asymptotically optimal path planner can solve the optimal planning problem asymptotically with the infinite number ($n \to \infty$) of samples,
\begin{equation}
        P\left(\lim_{n \to \infty}{\min_{\pi \in \Pi}{c(\pi)} = \pi^*}\right) = 1.
\end{equation}

\subsection{Adaptively Informed Trees}

This section reviews AIT* proposed in \cite{strub2020adaptively} and \cite{strub2021ait} briefly.
AIT* reduces the time cost of the most costly procedure: collision checking.
The adaptively informed heuristic uses the lazy-reverse search to guide the forward search.
With the help of the lazy-reverse search, the collision checking in the forward search only needs to check a small portion of the candidate edges.

The basic procedure of AIT* is shown in Algorithm \ref{AITAlgorithm}.
$sample(batchSize)$ takes a batch of $batchSize$ samples with the informed sampling method \cite{gammell2018informed} in the sampling stage.
$lazyReverseSearch()$ constructs a lazy-reverse search tree rooted at the goal.
The lazy-reverse tree is constructed lazily without collision checking, which means only vertices are valid and edges may be infeasible in the lazy-reverse tree.
$forwardSearch()$ incrementally constructs the forward tree whenever $forwardSearchMayImprove()$ returns true, which means the forward search may help to extend the forward tree towards the promising direction.
When the collision happens, $collide()$ returns true, and $updateLazyReverseSearch()$ reconstructs the related branches of the lazy-reverse tree.
AIT* uses $prune()$ to remove branches that can not help to improve the solution further.
At the end of each iteration, AIT* discards the whole lazy-reverse tree and reconstructs it from scratch in the next iteration.

\begin{algorithm}
    \caption{Adaptively Informed Trees}
    \label{AITAlgorithm}
    \While{\textbf{not} $terminate()$} {
        $sample(batchSize)$\;
        $lazyReverseSearch()$\;
        \While{$forwardSearchMayImprove()$}{
            $forwardSearch()$\;
            \If{$collide()$}{
                $updateLazyReverseSearch()$\;     \label{updateLazySearchingLineInAIT}
            }
        }
        $prune()$\;
    }
\end{algorithm}

\section{Methodology}

\renewcommand\arraystretch{1.7}
\begin{table*}[htbp]
    \centering
    \caption{The keys of $\mathcal{Q}_{FL}$, $\mathcal{Q}_{F}$, $\mathcal{Q}_{RL}$, $\mathcal{Q}_{R}$, $\mathcal{Q}_{meet}$.}
    \label{lexicographicalOrderKey}
    \begin{tabular}{ |p{3.5cm}<{\centering}|p{12.5cm}<{\centering}| } 
     \hline
     $x.key^{BiAIT*}_{\mathcal{Q}_{FL}}$ & $\lbrace \ \max(\min(x.\hat{h}_{g\text{-}F},\ x.\hat{h}_{rhs\text{-}F}) + x.\hat{h}_{E\text{-}g}, \ 2*\min(x.\hat{h}_{g\text{-}F},\ x.\hat{h}_{rhs\text{-}F})); \ \ \ \ \min(x.\hat{h}_{g\text{-}F},\ x.\hat{h}_{rhs\text{-}F}) \ \rbrace$ \\
     \hline
     $\lbrace x_p, x_c\rbrace.key^{BiAIT*}_{\mathcal{Q}_{F}}$ & $\lbrace \ x_p.g_F + \hat{c}(x_p, x_c) + x_c.\hat{h}_{g\text{-}R}; \ \ \ \ x_p.g_F + \hat{c}(x_p, x_c); \ \ \ \ x_p.g_F \ \rbrace$ \\ 
     \hline
     $x.key^{BiAIT*}_{\mathcal{Q}_{RL}}$ & $\lbrace \ \max(\min(x.\hat{h}_{g\text{-}R},\ x.\hat{h}_{rhs\text{-}R}) + x.\hat{h}_{E\text{-}s}, \ 2 * \min(x.\hat{h}_{g\text{-}R},\ x.\hat{h}_{rhs\text{-}R})); \ \ \ \ \min(x.\hat{h}_{g\text{-}R}, \ x.\hat{h}_{rhs\text{-}R}) \ \rbrace$ \\ 
     \hline
     $\lbrace x_p, x_c\rbrace.key^{BiAIT*}_{\mathcal{Q}_{R}}$ & $\lbrace \ x_p.g_R + \hat{c}(x_p, x_c) + x_c.\hat{h}_{g\text{-}F}; \ \ \ \ x_p.g_R + \hat{c}(x_p, x_c); \ \ \ \ x_p.g_R \ \rbrace$ \\ 
     \hline
     $\lbrace x_1, x_2\rbrace. key^{BiAIT*}_{\mathcal{Q}_{meet}}$ & $\lbrace \ x_1.g_F + c(x_1, x_2) + x_2.g_R \ \rbrace$ \\ 
     \hline
    \end{tabular}
    \vspace{-0.2cm}
\end{table*}

\subsection{Notation}

The valid state is denoted by $x \in \mathcal{X}_{free}$.
Let $\mathcal{X}_{s}$ stand for a set of valid samples.
$V$ represents the set of vertices, and each vertex in $V$ is associated with a valid state.
The directed edge is denoted by $\lbrace x_p, x_c \rbrace$, where the $x_p$ is the parent vertex and the $x_c$ is the child vertex.
Let $E$ be the set of edges, only edges in the forward and reverse trees are collision-free, and edges in lazy-forward and lazy-reverse trees may collide with $\mathcal{X}_{obs}$.
The forward tree, lazy-forward tree, reverse tree, and lazy-reverse tree are denoted by $\mathcal{T}_{F}$, $\mathcal{T}_{FL}$, $\mathcal{T}_{R}$, and $\mathcal{T}_{RL}$, respectively.
Each tree consists of vertices $V$ and edges $E$.
$\mathcal{T}_{F}$ and $\mathcal{T}_{FL}$ are rooted at $x_s$, $\mathcal{T}_{R}$ and $\mathcal{T}_{RL}$ are rooted at $x_{g}$.
LPA* \cite{koenig2004lifelong} uses $rhs$-value to refer to one-step lookahead cost of state $x$, and $g$-value to refer to the cost-from-root iff $x$ is locally consistent.
Our method uses $rhs$-value and $g$-value to denote the heuristic in lazy search processes.
The heuristic of state $x$ in the lazy-forward and lazy-reverse searches are denoted by $x.\hat{h}_{rhs\text{-}F}$, $x.\hat{h}_{g\text{-}F}$, $x.\hat{h}_{rhs\text{-}R}$, and $x.\hat{h}_{g\text{-}R}$, where the $F$ and $R$ in the subscripts represent forward and reverse, and $rhs$ and $g$ refer to $rhs$-value and $g$-value.
$x.\hat{h}_{rhs\text{-}F}$ and $x.\hat{h}_{rhs\text{-}R}$ denote the one-step lookahead heuristic of $x$ in the lazy-forward and the lazy-reverse searches, respectively.
$x.\hat{h}_{g\text{-}F}$ and $x.\hat{h}_{g\text{-}R}$ denote the cost-from-start heuristic via $\mathcal{T}_{FL}$ and cost-from-goal heuristic via $\mathcal{T}_{RL}$ iff $x$ is locally consistent.
The $rhs$-values of the root vertices of $\mathcal{T}_{FL}$ and $\mathcal{T}_{RL}$ are zero.
Let the neighbors of a vertex $x_i$ be $x_i._{neighbors}$, where $x_i._{neighbors}$ is a collection of vertices within a radius $r(q)$ of $x_i$.
$r(q)$ is defined as (\ref{Equation_NearNeighbors}) shows, where $q$ is the number of samples within the informed region, $\eta$, $n$, $\lambda\left(\mathcal{X}_{\hat{f}}\right)$, and $\zeta_n$ are the tuning parameters, dimensionality, and the Lebesgue measures of the informed set and an $n$-dimensional unit ball, respectively. 
\begin{equation}
    r(q) = \eta\left(2\left(1 + \frac{1}{n}\right)\left(\frac{\lambda\left(\mathcal{X}_{\hat{f}}\right)}{\zeta_n}\right)\left(\frac{\log\left(q\right)}{q}\right)\right)^\frac{1}{n}.
    \label{Equation_NearNeighbors}
\end{equation}

And the $rhs$-value of vertex $x_i$ is the minimum of heuristic edge cost between $x_i$ and $x_n \in x_i._{neighbors}$ plus the $g$-value of $x_{n}$:
\begin{equation}
\begin{split}
    & \hat{h}_{rhs\text{-}F}(x_i) = min\{x_n.\hat{h}_{g\text{-}F} + \hat{c}(x_n, x_i) | x_n \in x_i._{neighbors}\}; \\
    & \hat{h}_{rhs\text{-}R}(x_i) = min\{x_n.\hat{h}_{g\text{-}R} + \hat{c}(x_n, x_i) | x_n \in x_i._{neighbors}\}. 
    \label{updateRule}
\end{split}
\end{equation}

Vertex $x$ is deemed locally consistent in the lazy-forward search when $x.\hat{h}_{rhs\text{-}F}=x.\hat{h}_{g\text{-}F}$ and locally consistent in the lazy-reverse search when $x.\hat{h}_{rhs\text{-}R} = x.\hat{h}_{g\text{-}R}$.
The true cost-to-come from start via $\mathcal{T}_{F}$ is denoted by $x.g_F$, and the true cost-to-go from goal via $\mathcal{T}_{R}$ is denoted by $x.g_R$.
$x.\hat{h}_{E\text{-}s}$ and $x.\hat{h}_{E\text{-}g}$ denote the straight-line Euclidean distances to start and goal, respectively, using the front-to-end heuristic definition.
The heuristic cost and true connection cost between two states $x_1$ and $x_2$ is denoted by $\hat{c}(x_1, x_2)$ and $c(x_1, x_2)$, respectively.
Priority queues in BiAIT* contain multiple keys, and the queues are sorted in lexicographically ascending order.
In the sorting procedure, the elements with keys $a=a_{1}a_{2} \dots a_{k}$ and $b=b_{1}b_{2} \dots b_k$ in the queue are compared on the alphabetic order of the symbols in the first place $i$ where $a_i$ and $b_i$ differ (counting from the beginning of the keys): $a < b$ if and only if $a_i < b_i$.
Let $\mathcal{Q}$ be the priority queue sorted in lexicographically ascending order. 
AIT* defines the reverse-vertex queue $\mathcal{Q}^{AIT*}_{R}$ and the forward-edge queue $\mathcal{Q}^{AIT*}_{F}$ \cite{strub2021ait}.
Our method extends the queues in AIT* to four different priority queues: $\mathcal{Q}_{FL}$, $\mathcal{Q}_{F}$, $\mathcal{Q}_{RL}$, and $\mathcal{Q}_{R}$, where $\mathcal{Q}_{FL}$ and $\mathcal{Q}_{RL}$ denote the vertex-queues for lazy-forward/reverse search, $\mathcal{Q}_{F}$ and $\mathcal{Q}_{R}$ denote edge-queues for forward/reverse search.
Connection edges between $\mathcal{T}_{FL}$ and $\mathcal{T}_{RL}$ are stored in a set $\mathcal{S}_{lazyMeet}$, and connection edges between $\mathcal{T}_{F}$ and $\mathcal{T}_{R}$ are stored in a queue $\mathcal{Q}_{meet}$.
The queues $\mathcal{Q}_{FL}$, $\mathcal{Q}_{F}$, $\mathcal{Q}_{RL}$, $\mathcal{Q}_{R}$, and $\mathcal{Q}_{meet}$ are sorted in lexicographical order, and their keys are defined as Table \ref{lexicographicalOrderKey} shows.
Let $c_{cur}$ be the current solution cost of the planning problem.


\subsection{High-level Description}

Algorithm \ref{mainFunction} shows the bidirectional symmetrical planning process of BiAIT*.
At the very beginning, BiAIT* sets $c_{cur}$ to infinity. 
The start $x_s$ and the goal $x_g \in \mathcal{X}_{g}$ are added to $\mathcal{T}_{F}$ and $\mathcal{T}_{R}$, respectively.
When $lazySearchTerminate()$ returns true, candidate edges in $\mathcal{Q}_{F}$ may improve the solution, or the heuristic will not be better if the lazy search continues.
Otherwise, $lazySearch()$ will construct $\mathcal{T}_{FL}$ and provide the adaptive heuristic for the forward and reverse searches.
When the best edge in $\mathcal{Q}_{F}$ may improve the current solution, $forwardSearchMayImprove()$ will return true, and $\mathcal{T}_{F}$ will grow according to the heuristic.
If the lazy search and forward/reverse search processes terminate, BiAIT* will dispose of $\mathcal{T}_{FL}$ and $\mathcal{T}_{RL}$, prune $\mathcal{X}_{s}$, $\mathcal{T}_{F}$, and $\mathcal{T}_{R}$, and take a new batch of samples in the informed space.
We utilize the same pruning and sampling strategy (Algorithm \ref{mainFunction}, Line \ref{pruneLine} and \ref{sampleLine}) as AIT* \cite{strub2021ait}, details can be found in the Appendix.
$initBatch()$ initializes $\mathcal{X}_s$ after sampling.
$swap()$ switches $\mathcal{T}_{F}, \mathcal{T}_{FL}, \mathcal{Q}_{F}, \mathcal{Q}_{FL}$ and $\mathcal{T}_{R}, \mathcal{T}_{RL}, \mathcal{Q}_{R}, \mathcal{Q}_{RL}$, which enables BiAIT* to search along the forward and reverse direction alternately.
We introduce the details of our method as follows.

\begin{algorithm}
    \caption{Bidirectional AIT*}
    \label{mainFunction}
    $c_{cur} \gets \infty$; \ \ 
    $\mathcal{T}_{F}.insert(x_{s})$; \ \  $\mathcal{T}_{R}.insert(x_{g}$)\;
    \While{\textbf{not} $terminate()$ }{
        \If(){\textbf{not} $lazySearchTerminate()$}{ \label{lazySearchTerminateFunction}
            $lazySearch()$\; \label{lazySearchFunction}
        }
        \ElseIf{$forwardSearchMayImprove()$}{  \label{validSearchMayImproveCall}
            $forwardSearch()$\; \label{mainFunction_ForwardSearch}
        }
        \Else{
            $prune(\mathcal{X}_{s}, \mathcal{T}_{F}, \mathcal{T}_{R})$\;   \label{pruneLine}
            $\mathcal{X}_{s} \gets \mathcal{X}_{s} \cup $ $sample(batchSize$, $c_{cur})$\;  \label{sampleLine}
            $initBatch()$\;   \label{initBatchFunction}
        }
        $swap($\{$\mathcal{T}_{F}$, $\mathcal{T}_{FL}$, $\mathcal{Q}_{F}$, $\mathcal{Q}_{FL}$\}, \{$\mathcal{T}_{R}$, $\mathcal{T}_{RL}$, $\mathcal{Q}_{R}$, $\mathcal{Q}_{RL}$\}$)$\;
    }
\end{algorithm}

\subsection{Details}

\subsubsection{Initialize the batch}

BiAIT* initializes $\mathcal{X}_s$ and prepares for the lazy search (Algorithm \ref{mainFunction}, Line \ref{initBatchFunction}) after adding a new batch of samples.
Algorithm \ref{initBatchAlgorithm} describes the process of the batch initialization.
We reconstruct $\mathcal{T}_{FL}$ and $\mathcal{T}_{RL}$ from scratch; the $rhs$-value heuristic and $g$-value heuristic in the previous lazy search are set to infinite. 
To fully take advantage of the information of the previous forward and reverse searches, $\mathcal{T}_{F}$ and $\mathcal{T}_{R}$ are copied to the corresponding new lazy trees (Algorithm \ref{initBatchAlgorithm}, Line \ref{copyStart} to \ref{copyEnd}). 
The $rhs$-value of $x_s$ in lazy-forward search ($x_s.h_{rhs\text{-}F}$) and the $rhs$-value of goal states in lazy-reverse search ($x_g.h_{rhs\text{-}R}$) are set to zero.
The start and the goal are added to the corresponding lazy queues.
$expand()$ in Algorithm \ref{initBatchAlgorithm}, Line \ref{expandLine} is same as that of AIT* \cite{strub2021ait}, whose details are shown in the Appendix.

\begin{algorithm}[t]
    \caption{$initBatch()$}
    \label{initBatchAlgorithm}
    $clear(\mathcal{T}_{FL}, \mathcal{T}_{RL}, \mathcal{Q}_{F}, \mathcal{Q}_{FL}, \mathcal{Q}_{R}, \mathcal{Q}_{RL}, \mathcal{S}_{lazyMeet})$\;
    \ForAll{$x \in \mathcal{X}_{s}$}{
        \If{$x \in \mathcal{T}_{F}$}{  \label{copyStart}
            $x.\hat{h}_{rhs\text{-}F} \gets x.g_F$; \ \ \ $\mathcal{Q}_{FL}.insert(x)$\;
        }
        \ElseIf{$x \in \mathcal{T}_{R}$}{
            $x.\hat{h}_{rhs\text{-}R} \gets x.g_R$; \ \ \ $\mathcal{Q}_{RL}.insert(x)$\; 
        }                               \label{copyEnd}
        \Else{
            $x.\hat{h}_{rhs\text{-}F} \gets \infty$; \ \ \ $x.\hat{h}_{g\text{-}F} \gets \infty$\; 
            $x.\hat{h}_{rhs\text{-}R} \gets \infty$; \ \ \ $x.\hat{h}_{g\text{-}F} \gets \infty$\; 
        }
    }
    $x_s.h_{rhs\text{-}F} \gets 0$; \ \ \ $x_g.h_{rhs\text{-}R} \gets 0$\;
    $\mathcal{Q}_{F}.insert(expand(x_{s}))$; \ \ \ $\mathcal{Q}_{R}.insert(expand(x_g))$\; \label{expandLine}
\end{algorithm}

\subsubsection{Lazy search}

The lazy search stops when the heuristic can not be improved or the forward and reverse search may have chance to improve $\mathcal{T}_{F}$ and $\mathcal{T}_{R}$ (Algorithm \ref{mainFunction}, Line \ref{lazySearchTerminateFunction}).
Details of $lazySearchTerminate()$ are shown in Algorithm \ref{AlgOtherFuncs}, Line \ref{FuncLazySearchTerminate} to \ref{FuncLazySearchTerminateEnd}.


The details of $lazySearch()$ (Algorithm \ref{mainFunction}, Line \ref{lazySearchFunction}) are described in Algorithm \ref{lazySearchAlgorithm}, which extends the $\mathcal{T}_{FL}$ gradually towards the goal. 
BiAIT* extracts the best candidate vertex $x$ in the forward vertex queue $\mathcal{Q}_{FL}$, which has the highest chance to improve the heuristic.
If the $rhs$-value of $x$ is less than the $g$-value of $x$, the $g$-value will be updated, and the vertex $x$ will become locally consistent.
Otherwise, the $g$-value will be set to infinity and the $updateState()$ (Algorithm \ref{lazySearchAlgorithm}, Line \ref{updateStateFunction_1}) will recalculate the $rhs$-value heuristic for $x$.
All vertices potentially affected by the change of $x$ are updated as well as the corresponding order in the priority vertex queue (Algorithm \ref{lazySearchAlgorithm}, Line \ref{updateStateNeighborsStart} to \ref{updateStateNeighborsEnd}).

\begin{algorithm}[t]
    \caption{$lazySearch()$}
    \label{lazySearchAlgorithm}
        $x \gets Q_{FL}.\textup{popBest()}$\;
        \If{$x.\hat{h}_{rhs\text{-}F} < x.\hat{h}_{g\text{-}F}$}{
            $x.\hat{h}_{g\text{-}F} \gets x.\hat{h}_{rhs\text{-}F}$\;
        }
        \Else{
            $x.\hat{h}_{g\text{-}F} \gets \infty$\;
            $updateState(x)$\; \label{updateStateFunction_1}
        }
        \ForAll{$x_{neighbor} \in x._{neighbors} \backslash x._{black}$}{   \label{updateStateNeighborsStart}
            $updateState(x_{neighbor})$\;   \label{updateStateFunction_2}
        }   \label{updateStateNeighborsEnd}
\end{algorithm}

Algorithm \ref{updateStateAlgorithm} describes the details of $updateState()$ (Algorithm \ref{lazySearchAlgorithm}, Line \ref{updateStateFunction_1} and \ref{updateStateFunction_2}).
If the vertex $x$ is not in the lazy-reverse or reverse trees (Algorithm \ref{updateStateAlgorithm}, Line \ref{notInRLTree}), BiAIT* will update the $rhs$-value and the lazy parent of $x$ (Algorithm \ref{updateStateAlgorithm}, Line \ref{findLazyParentStart} to Line \ref{findLazyParentEnd}).
The update rule of $rhs$-value follows (\ref{updateRule}).
If vertex $x$ is not locally consistent, the state $x$ will be inserted into $\mathcal{Q}_{FL}$ or the associated element in $\mathcal{Q}_{FL}$ will be updated with the new key of $x$.
Otherwise, we remove it from $\mathcal{Q}_{FL}$.
Algorithm \ref{lazySearchAlgorithm} and \ref{updateStateAlgorithm} is a bidirectional version of $ComputeShortestPath()$ and $UpdateVertex()$ in LPA* \cite{koenig2004lifelong} in continuous space.
If the vertex $x$ is in the lazy-reverse or reverse trees (Algorithm \ref{updateStateAlgorithm}, Line \ref{inRLTree}), we search the neightbors $x_i \in \mathcal{T}_{FL} \cup \mathcal{T}_{F}$ of state $x$.
When $x$ is consistent in the lazy-reverse search, and $x_i$ is consistent in the lazy-forward search, we connect the $\mathcal{T}_{FL}$ and $\mathcal{T}_{RL}$ with a lazy meet edge $\lbrace x_i, x \rbrace$.
The lazy meet edges are stored in a set $\mathcal{S}_{lazyMeet}$ (Algorithm \ref{updateStateAlgorithm}, Line \ref{storeLazyConnectionLine}).
$\mathcal{T}_{FL}$ and $\mathcal{T}_{RL}$ will share the heuristic information and update the queues after establishing the connection (Algorithm \ref{updateStateAlgorithm}, Line \ref{lazyTreeMeetFunction}).

\begin{algorithm}[t]
    \caption{$updateState(x)$}
    \label{updateStateAlgorithm}
    \If{$x \neq x_{s}$ \textbf{and} $x \notin \mathcal{T}_{RL}$ \textbf{and} $x \notin \mathcal{T}_{R}$}{    \label{notInRLTree}
        $x_{p} \gets \arg \min_{x_i \in x._{neighbors} \backslash x._{black}} \lbrace x_i.\hat{h}_{g} + \hat{c}(x, x_i) \rbrace$\; \label{findLazyParentStart}
        $x.\hat{h}_{rhs} \gets x_p.\hat{h}_{g} + \hat{c}(x, x_p)$\;
        $x._{lazyParent}$ $\gets$ $x_p$; \ $x_c._{lazyChildren}.insert(x_p)$\;  \label{findLazyParentEnd}
        \If{\textbf{not} $consistent(x)$}{
            \If{$x \notin Q_{L}$}{
                $Q_L$.$insert(x)$\;
            }
            \ElseIf{$x \in Q_{L}$}{
                $Q_L$.$update(x)$\;
            }
        } 
        \Else{
            $Q_L$.$erase(x)$\;
        }
    }
    \ElseIf{$x \in \mathcal{T}_{RL}$ \textbf{or} $x \in \mathcal{T}_{R}$}{     \label{inRLTree}
        \ForAll{$x_{i} \in x_.{neighbors}$ \textbf{and} $x_{i} \in \mathcal{T}_{FL} \cup \mathcal{T}_{F}$}{
            \If{$consistent(x)$ \textbf{and} $consistent(x_i)$}{
                $\mathcal{S}_{lazyMeet}$.$insertOrUpdate(\lbrace x_i, x \rbrace)$\; \label{storeLazyConnectionLine}
                $lazyTreesMeet(\lbrace x_i, x \rbrace)$\; \label{lazyTreeMeetFunction}
            }
        }   
    }
\end{algorithm}

$lazyTreesMeet()$ (Algorithm \ref{updateStateAlgorithm}, Line \ref{lazyTreeMeetFunction}) propagates the heuristic along with the lazy trees, which is illustrated by Algorithm \ref{lazyTreesMeetAlgorithm}.
With a little counterintuitive, the forward search is guided by the lazy-reverse search, and the reverse search is guided by the lazy-forward search.
Therefore, Algorithm \ref{lazyTreesMeetAlgorithm} propagates the $x.\hat{h}_{g\textup{-}R}$ in $\mathcal{T}_{FL}$ and $x.\hat{h}_{g\textup{-}F}$ in $\mathcal{T}_{RL}$.
BiAIT* also updates affected elements in the forward and reverse edge queues.
We connect the $\mathcal{T}_{FL}$ and $\mathcal{T}_{RL}$ greedily without considering whether the new connection is better than the existing connections.

\begin{algorithm}[t]
    \caption{$lazyTreesMeet(\lbrace x_1, x_2 \rbrace)$}
    \label{lazyTreesMeetAlgorithm}
    \If{$x_2.\hat{h}_{g\textup{-}R} + \hat{c}(x_1, x_2) < x_1.\hat{h}_{g\textup{-}R}$ \textbf{and} $x_1 \in \mathcal{T}_{FL}$}{
        $x_1.\hat{h}_{g\textup{-}R} \gets x_2.\hat{h}_{g\textup{-}R} + \hat{c}(x_1, x_2)$\;
        $lazyTreesMeet(x_1._{lazyParent}, x_1)$\;
        \If{$\mathcal{Q}_{F}.contain(\lbrace \_, x_1 \rbrace)$}{
            $\mathcal{Q}_{F}.update(\lbrace \_, x_1 \rbrace)$\;
        }
    }
    \If{$x_1.\hat{h}_{g\textup{-}F} + \hat{c}(x_2, x_1) < x_2.\hat{h}_{g\textup{-}F}$ \textbf{and} $x_2 \in \mathcal{T}_{RL}$}{
        $x_2.\hat{h}_{g\textup{-}F} \gets x_1.\hat{h}_{g\textup{-}F} + \hat{c}(x_2, x_1)$\;
        $lazyTreesMeet(x_2._{lazyParent}, x_2)$\;
        \If{$\mathcal{Q}_{R}.contain(\lbrace \_, x_2 \rbrace)$}{
            $\mathcal{Q}_{R}.update(\lbrace \_, x_2 \rbrace)$\;
        }
    }
\end{algorithm}

\begin{algorithm}[t]
    \caption{$forwardSearch()$}
    \label{validSearchAlgorithm}
    $\lbrace x_p, x_c \rbrace \gets Q_{F}.popBest()$\; 
    \If{$\lbrace x_p, x_c \rbrace \in \mathcal{T}_{F}$}{
        $\mathcal{Q}_{F}.insert(expand(x_c)$)\;
    }
    \ElseIf{$x_p.g_F + \hat{c}(x_p, x_c) < x_c.g_F$}{
        \If{$x_c \in x_p._{white}$ \textbf{or} $collisionFree(\lbrace x_p, x_c \rbrace)$}{      \label{LineCollisionChecking}
            $x_p._{white}.insert(x_c)$; \ \ $x_c._{white}.insert(x_p)$\;            \label{LineMemoryWhiteList}
            \If{$x_p.g_F + c(x_p, x_c) + x_c.\hat{h}_{g\text{-}R} < c_{cur}$}{   \label{canImproveCurrentSolutionLine}
                \If{$x_c \in \mathcal{T}_{R}$}{        \label{inRVTreeLine}
                    $\mathcal{Q}_{meet} \gets \lbrace x_p, x_c \rbrace$\;  \label{insertInToValidMeetQueueLine}
                    $updateSolution()$;           \label{updateSolutionLine}
                }
                \ElseIf{$x_p.g_F + c(x_p, x_c) < x_c.g_F$}{       \label{canImproveForwardTreeLine}
                    $x_c._{parent} \gets x_p$\;
                    $x_p._{children}.insert(x_p)$\;
                    $\mathcal{T}_{F}.insert(\lbrace x_p, x_c \rbrace)$\;     \label{expandAndRewireLine}
                    $\mathcal{Q}_{F}.insert(expand(x_c))$\;
                }
            }
        }
        
        \ElseIf{$\lbrace x_p, x_c \rbrace \in \mathcal{T}_{FL} \cup \mathcal{S}_{lazyMeet} \cup \mathcal{T}_{RL}$}{
            $x_p._{black}.insert(x_c)$; \ \ $x_c._{black}.insert(x_p)$\;        \label{LineMemoryBlackList}
            $updateLazySearch(\lbrace x_p, x_c \rbrace)$\; \label{updateLazySearchLine}
        }
    }
\end{algorithm}

\subsubsection{Forward and reverse searches}

When the function $forwardSearchMayImprove()$ (Algorithm \ref{mainFunction}, Line \ref{validSearchMayImproveCall} and Algorithm \ref{AlgOtherFuncs}, Line \ref{FuncForwardSearchMayImprove}) returns true, BiAIT* will try to construct the $\mathcal{T}_{F}$ with the most promising candidate edge.
Details of $forwardSearch()$ is presented in Algorithm \ref{validSearchAlgorithm}.
The best candidate edge $\lbrace x_p, x_c \rbrace$ in $\mathcal{Q}_{F}$ is extracted, where $x_p$ and $x_c$ are the parent and the child of the edge.
If the connection between $x_c$ and $x_p$ has already been collision-checked ($x_c \in x_p._{white}$), BiAIT* will treat $\lbrace x_p, x_c \rbrace$ as feasible edge, otherwise, connection validator will check whether $\lbrace x_p, x_c \rbrace$ collides with $\mathcal{X}_{obs}$.
The black and white lists ($x._{black}$ and $x._{white}$) can memory the result of the collision checking (Algorithm \ref{validSearchAlgorithm}, Line \ref{LineMemoryWhiteList} and \ref{LineMemoryBlackList}) to avoid processing the edge repeatedly.
If the candidate edge can improve the current solution (Algorithm \ref{validSearchAlgorithm}, Line \ref{canImproveCurrentSolutionLine}) and $x_c$ is in $\mathcal{T}_{R}$ (Algorithm \ref{validSearchAlgorithm}, Line \ref{inRVTreeLine}), BiAIT* will connect $\mathcal{T}_{F}$ and $\mathcal{T}_{R}$ and update the solution cost $c_{cur}$ (Algorithm \ref{validSearchAlgorithm}, Line \ref{insertInToValidMeetQueueLine} and \ref{updateSolutionLine}).
$updateSolution()$ (Algorithm \ref{validSearchAlgorithm}, Line \ref{updateSolutionLine}) extracts the best edge $\lbrace x^{best}_p, x^{best}_c \rbrace$ in $\mathcal{Q}_{meet}$, and updates $c_{cur}$.
The solution path is the union of the inverse path of tracing from $x^{best}_p$ to $x_s$ via $\mathcal{T}_{F}$ and the path tracing from $x^{best}_c$ to $x_g$ via $\mathcal{T}_{R}$.
If the candidate edge can improve current solution (Algorithm \ref{validSearchAlgorithm}, Line \ref{canImproveCurrentSolutionLine}) and can improve $\mathcal{T}_{F}$ (Algorithm \ref{validSearchAlgorithm}, Line \ref{canImproveForwardTreeLine}), $\mathcal{T}_{F}$ will extend and rewire (Algorithm \ref{validSearchAlgorithm}, Line \ref{expandAndRewireLine}).
$updateLazySearch()$ (Algorithm \ref{validSearchAlgorithm}, Line \ref{updateLazySearchLine}) will update relevant lazy branches when the collision happens. 
The details of $updateLazySearch()$ is presented by Algorithm \ref{updateLazySearchAlgorithm}.


\begin{algorithm}[t]
    \caption{$updateLazySearch(\lbrace x_p, x_c \rbrace)$} 
    \label{updateLazySearchAlgorithm}
    $x_c.h_{rhs\text{-}F} \gets \infty$; \ \ \ $x_c.h_{g\text{-}F} \gets \infty$\;    \label{updateLazySearchToInfLine}
    $x_p._{lazyChildren}.remove(x_c)$\; \label{removeLazyChildrenLine}
    $x_c._{lazyParent}.clear()$\;   \label{removeLazyParentLine}

    \If{$x_c \in \mathcal{S}_{lazyMeet}$}{
        \ForAll{$\lbrace x_c, x' \rbrace \in \mathcal{S}_{lazyMeet}.contain(x_c)$}{
            $x_c.\hat{h}_{g\text{-}R} \gets \infty \ \ \ x'.\hat{h}_{g\text{-}F} \gets \infty$\;
            $\mathcal{S}_{lazyMeet}.remove(\lbrace x_c, x' \rbrace)$\;
            \ForAll(// All lazy edges end with $x'$.){$\lbrace x'', x' \rbrace \in \mathcal{S}_{lazyMeet}$}{
                $updatePredecessorToStart(x'')$\;    \label{lineUpdatePredecessorLineToStart}
                $updatePredecessorToGoal(x')$\;    \label{lineUpdatePredecessorToGoal}
            }
        }
    }
    \ForAll{$x_i \in x_c._{lazyChildren}$}{
        $updateLazySearch(\lbrace x_c, x_i \rbrace)$\;
    }
    $updateState(x_c)$\;
\end{algorithm}

\begin{algorithm}[t]
    \caption{Other Functions}
    \label{AlgOtherFuncs}
    \SetKwFunction{forwardSearchMayImprove}{$forwardSearchMayImprove$}
    \SetKwFunction{lazySearchTerminate}{$lazySearchTerminate$}
    \SetKwFunction{updatePredecessorToStart}{$updatePredecessorToStart$}
    \SetKwFunction{updatePredecessorToGoal}{$updatePredecessorToGoal$}
    \SetKwProg{CustomFunc}{Function}{}{}

    \CustomFunc{\lazySearchTerminate$()${}}{        \label{FuncLazySearchTerminate}
        \If{$\mathcal{Q}_F.empty()$ \textbf{or} $\mathcal{Q}_R.empty()$ 
            \textbf{or} 
            $\mathcal{Q}_{FL}.empty()$}{
            \Return $true$\;
        }
        $bestEdge \gets better(\mathcal{Q}_F.best(), \mathcal{Q}_R.best())$\;
        $x_{best} \gets \mathcal{Q}_{FL}.best()$\;
        \If{$\min(x_{best}.\hat{h}_{g\text{-}F},\ x_{best}.\hat{h}_{rhs\text{-}F}) + x_{best}.\hat{g}_{R} 
            > bestEdge.key^{BiAIT*}_{\mathcal{Q}_{F}}[0]$}{
            \Return $true$\;
        }
        \Else {
            \Return $false$\;           \label{FuncLazySearchTerminateEnd}
        }
    }

    \CustomFunc{\forwardSearchMayImprove$()${}}{        \label{FuncForwardSearchMayImprove}
        \Return $\mathcal{Q}_F.best().key^{BiAIT*}_{\mathcal{Q}_{F}}[0] < c_{cur}$
    }

    \CustomFunc{\updatePredecessorToStart$(x)${}}{      \label{LineUpdatePredecessorToStart}
        $x.\hat{h}_{g\text{-}R} \gets \infty$\;
        \ForAll{$x' \in x._{lazyChildren}$}{
            $x.\hat{h}_{g\text{-}R} \gets \min(x'.\hat{h}_{g\text{-}R} + \hat{c}(x', x), \ x.\hat{h}_{g\text{-}R})$\;
        }
        \If{$\mathcal{Q}_{F}.contain(\lbrace \_, x \rbrace)$}{
            $\mathcal{Q}_{F}.update(\lbrace \_, x \rbrace)$\;
        }
        \If{\textbf{not} $x.\hat{h}_{g\text{-}R} == \infty$} {
            $\updatePredecessorToStart(x._{lazyParent})$\;       \label{LineCallUpdateToStartRecursively}
        }
    }

    \CustomFunc{\updatePredecessorToGoal$(x)${}}{       \label{updatePredecessorToGoal}
        $x.\hat{h}_{g\text{-}F} \gets \infty$\;
        \ForAll{$x' \in x._{lazyChildren}$}{
            $x.\hat{h}_{g\text{-}F} \gets \min(x'.\hat{h}_{g\text{-}F} + \hat{c}(x', x), \ x.\hat{h}_{g\text{-}F})$\;
        }
        \If{$\mathcal{Q}_{R}.contain(\lbrace \_, x \rbrace)$}{
            $\mathcal{Q}_{R}.update(\lbrace \_, x \rbrace)$\;
        }
        \If{\textbf{not} $x.\hat{h}_{g\text{-}F} == \infty$} {
            $\updatePredecessorToGoal(x._{lazyParent})$\;       \label{LineCallUpdateToGoalRecursively}
        }
    }
\end{algorithm}

$updateLazySearch()$ (Algorithm \ref{updateLazySearchAlgorithm}) executes recursively and propagates the changed heuristic.
When calling $updateLazySearch()$, $rhs$-value and $g$-value of $x_c$ are set to infinity (Algorithm \ref{updateLazySearchAlgorithm}, Line \ref{updateLazySearchToInfLine}),
and the lazy link between $x_p$ and $x_c$ will be broken (Algorithm \ref{updateLazySearchAlgorithm}, Line \ref{removeLazyChildrenLine} and \ref{removeLazyParentLine}).
If the vertex $x_c$ is a vertex connects $\mathcal{T}_{FL}$ and $\mathcal{T}_{RL}$, the function $updatePredecessorToStart(x)$ and $updatePredecessorToGoal(x)$ will be called (Algorithm \ref{updateLazySearchAlgorithm}, Line \ref{lineUpdatePredecessorLineToStart} and \ref{lineUpdatePredecessorToGoal}).
Details of $updatePredecessorToStart(x)$ and $updatePredecessorToGoal(x)$ are shown in Algorithm. \ref{AlgOtherFuncs}, Line \ref{LineUpdatePredecessorToStart} to \ref{LineCallUpdateToGoalRecursively}.

\begin{figure}[t]
    \vspace{-0.2cm}
    \flushright
    \includegraphics[width=0.95\linewidth]{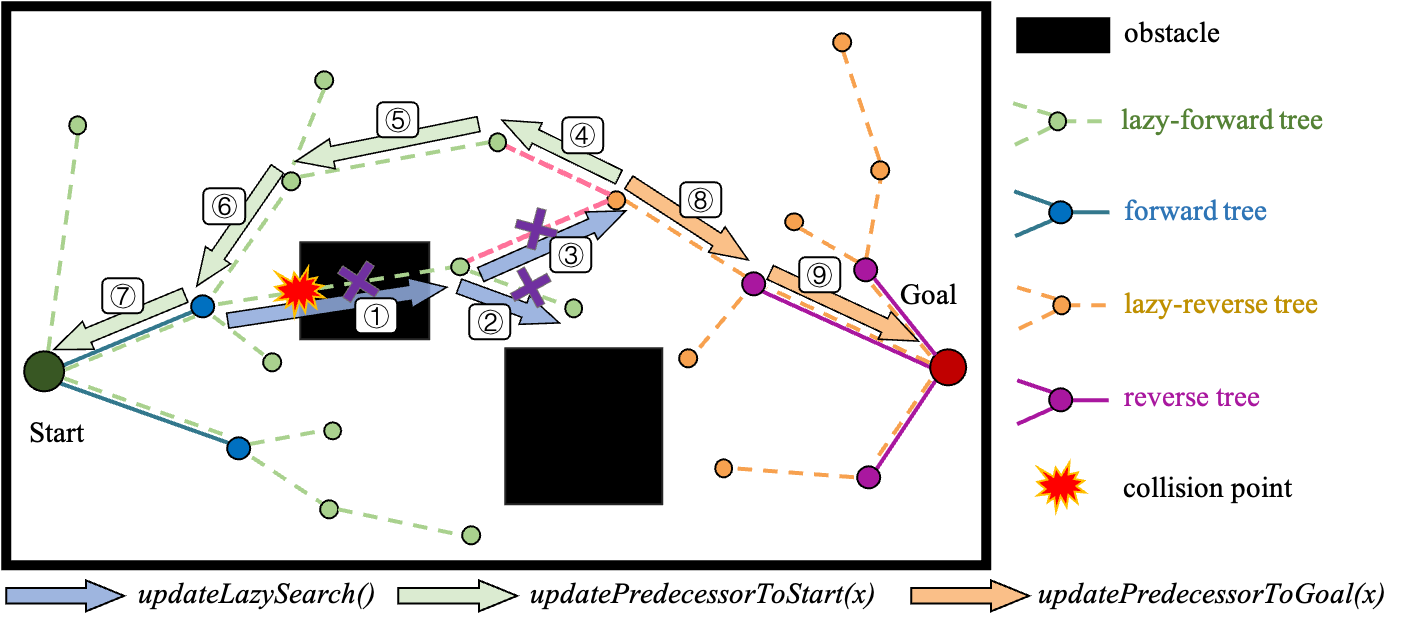}
    
    \caption{
    The update processes after the collision happend.
    The order that edges are processed is indicated by the numbers {\small\ding{192}} to {\small\ding{200}}.
    Blue arrows show the propagation path of $updateLazySearch(\lbrace x_p, x_c \rbrace)$, where all lazy edges processed by it will be invalidated.
    Yellow and green arrows show the propagation path of $updatePredecessorToStart(x)$ and $updatePredecessorToGoal(x)$, respectively, which do not influence the struture of $\mathcal{T}_{FL}$ and $\mathcal{T}_{RL}$.
    }
    \label{diffInInvalidation}
    \vspace{-0.2cm}
\end{figure}

In $lazySearchTerminate()$ (Algorithm \ref{AlgOtherFuncs}, Line \ref{FuncLazySearchTerminate}), BiAIT* terminates the lazy search if any one of $\mathcal{Q}_{FL}$, $\mathcal{Q}_{F}$, or $\mathcal{Q}_{R}$ is empty. 
The lazy search also terminates when the key of the best edge in the forward and reverse edge queues ($\mathcal{Q}_{F}$ and $\mathcal{Q}_{R}$) is better than the key of the best vertex in the forward vertex queue ($\mathcal{Q}_{FL}$), which means the forward or reverse search can possibly improve the current solution. 
$updatePredecessorToStart(x)$ (Algorithm \ref{AlgOtherFuncs}, Line \ref{LineUpdatePredecessorToStart} to \ref{LineCallUpdateToStartRecursively}) and $updatePredecessorToGoal(x)$ (Algorithm \ref{AlgOtherFuncs}, Line \ref{updatePredecessorToGoal} to \ref{LineCallUpdateToGoalRecursively}) check whether the heuristic is propagated from the invalidated branch; if yes, BiAIT* will update the heuristic, $\mathcal{Q}_{F}$, and $\mathcal{Q}_{R}$. 
Fig. \ref{diffInInvalidation} illustrates the differences between these three functions that are called after a collision occurs.
Among them, only $updateLazySearch(\lbrace x_p, x_c \rbrace)$ influences the structure of $\mathcal{T}_{FL}$ and $\mathcal{T}_{RL}$.
When updating the heuristic with $updatePredecessorToStart(x)$ and $updatePredecessorToGoal(x)$, the structures of $\mathcal{T}_{FL}$ and $\mathcal{T}_{RL}$ are well maintained.

\section{Simulations}

\begin{figure*}[htbp]
    \centering
    \begin{minipage}{0.25\textwidth}
        \captionsetup[subfloat]{labelformat=empty}
        \subfloat[{\footnotesize BugTrap}]{\includegraphics[width=0.98\linewidth]{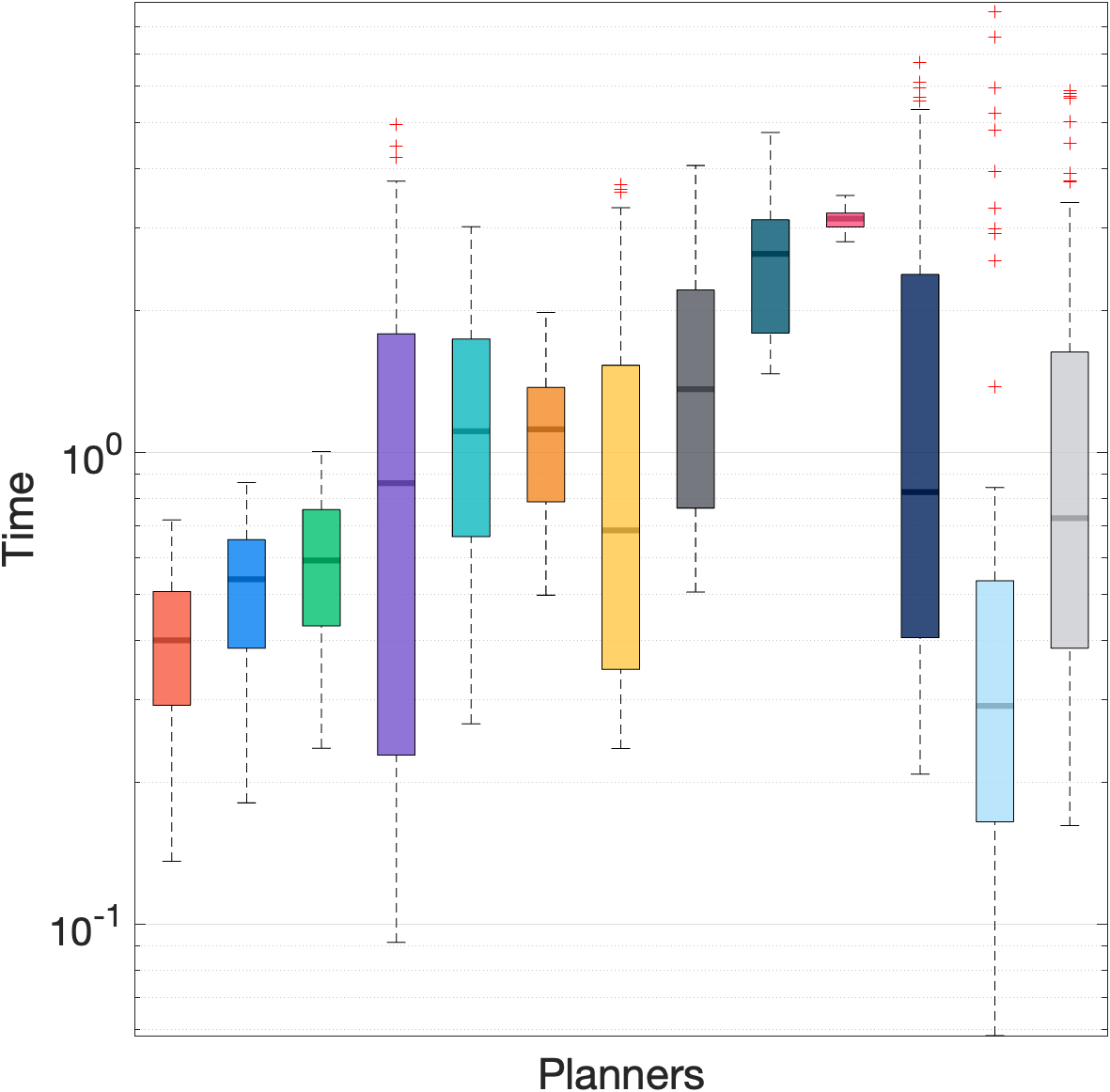}}%
    \end{minipage}
    \hfill
    \begin{minipage}{0.25\textwidth}
        \captionsetup[subfloat]{labelformat=empty}
        \subfloat[{\footnotesize Maze}]{\includegraphics[width=0.98\linewidth]{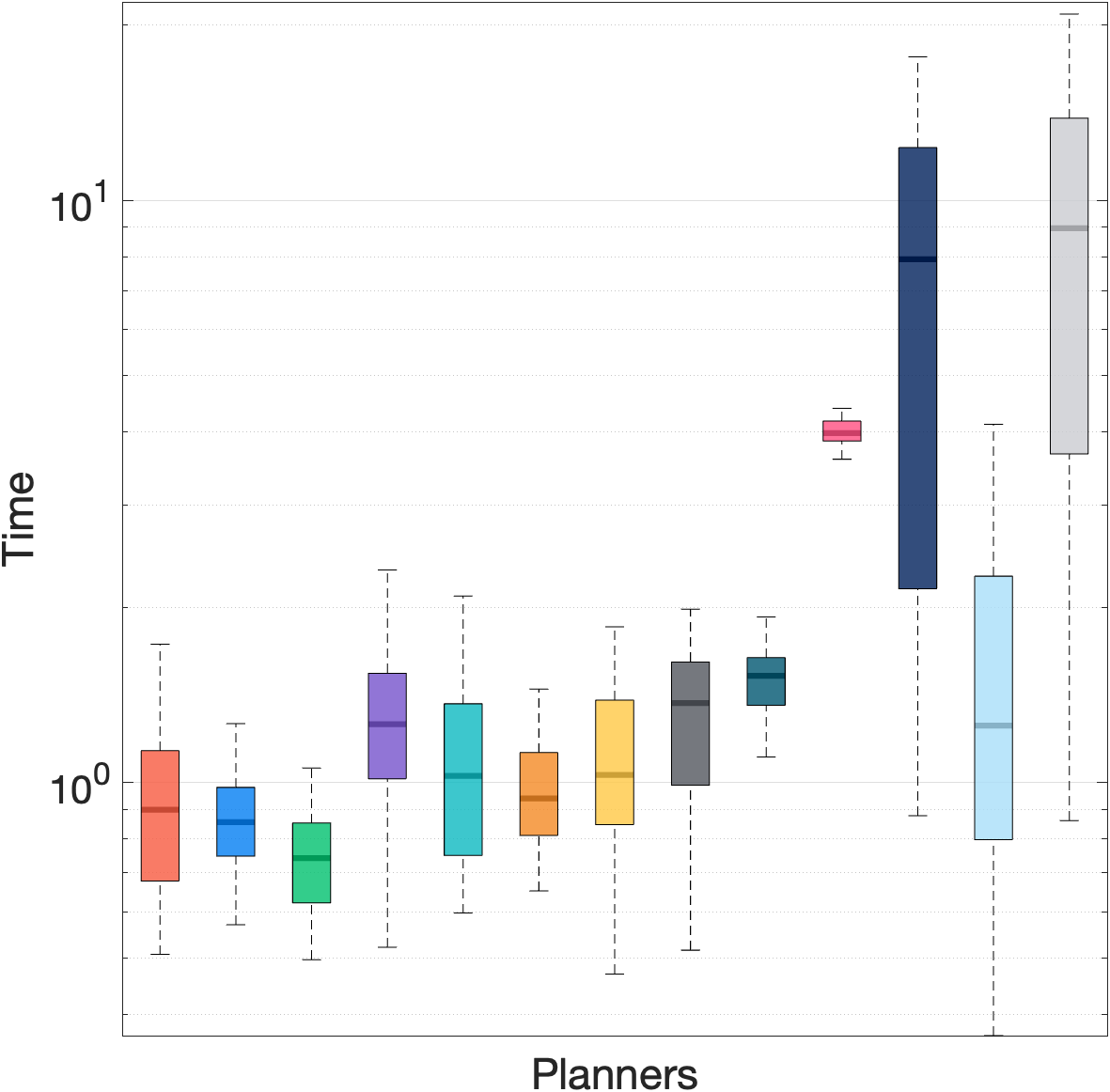}}%
    \end{minipage}
    \hfill
    \begin{minipage}{0.25\textwidth}
        \captionsetup[subfloat]{labelformat=empty}
        \subfloat[{\footnotesize Easy(3D)}]{\includegraphics[width=0.98\linewidth]{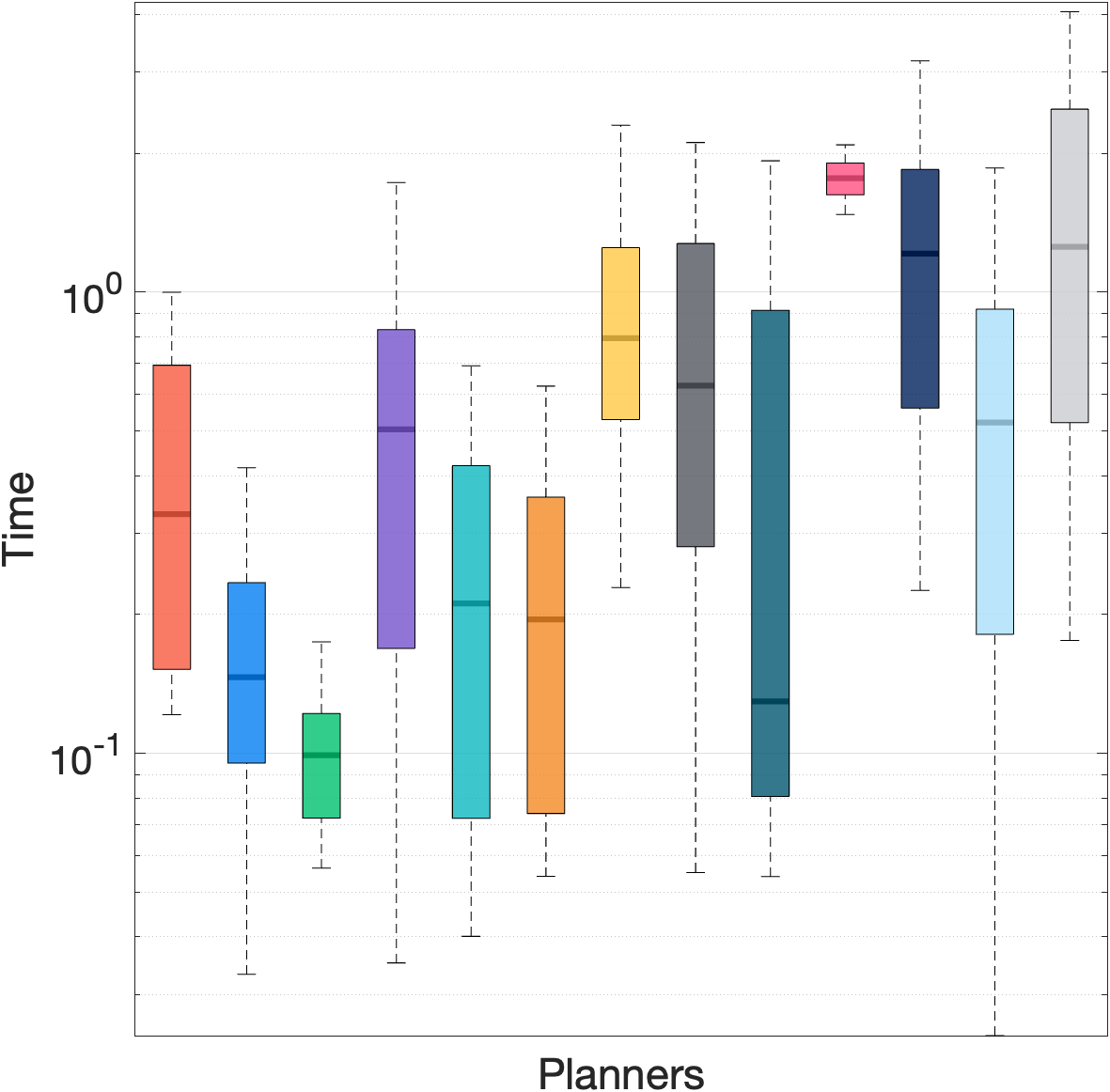}}%
    \end{minipage}
    \hfill
    \begin{minipage}{0.11\textwidth}
        \captionsetup[subfloat]{labelformat=empty}
        \subfloat[]{\includegraphics[width=0.98\linewidth]{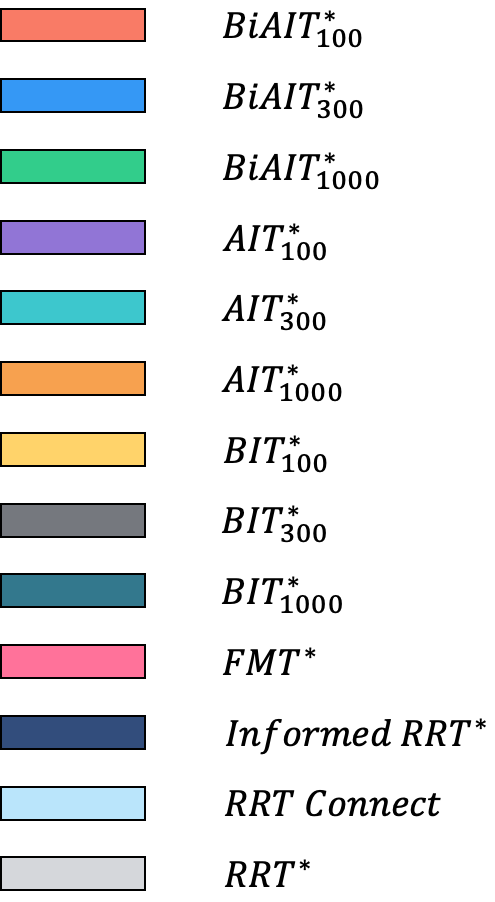}}%
    \end{minipage}
    \caption{The computational time consumed by each planner of finding the initial feasible solution in OMPL $SE(2)$ and $SE(3)$ environments. The central solid lines and cross points are the median values and outliers. The bottom and top edges of the boxes indicate the $25\%$ and $75\%$, respectively. The whiskers show extreme values except for the outliers. The base-10 logarithmic scale is used on the vertical axis. The collision checking interval is $0.001$. All planners are implemented with C++, and simulations are conducted on a laptop with Apple Silicon M1 Pro CPU. }
    \label{OMPLSimResult}
    \vspace{-0.2cm}
\end{figure*}

\subsection{Simulations with OMPL}

The Open Motion Planning Library (OMPL) \cite{sucan2012open} includes various well-implemented planning algorithms and a benchmark platform.
Therefore, we evaluate BiAIT* with the OMPL benchmark platform and compare it against several planners in OMPL.
The comparison includes AIT*, BIT*, FMT*, Informed RRT*, RRT-Connect, and RRT*.
Three OMPL benchmark environments are considered, the `BugTrap' and `Maze' environments in $SE(2)$ space as well as `Easy(3D)' in $SE(3)$ space.
In the `BugTrap' environment, the start is set inside the bugtrap and the goal is set outside.
The robot in the simulations is a union of several polyhedrons.

\begin{table*}[htbp]
    \centering
    \caption{Initial solution qualities of BiAIT*, AIT*, BIT*, FMT*, Informed RRT*, RRT-Connect, and RRT*. Subscripts show $batchSize$ in BiAIT*, AIT*, and BIT*}
    \label{InitialSolutionQualityTable}
    \resizebox{\textwidth}{10mm}{
    \begin{tabular}{|c | c c c c c c c c c c c c c|} 
     \hline
     & BiAIT*$_{100}$ & BiAIT*$_{300}$ & BiAIT*$_{1000}$ & AIT*$_{100}$ & AIT*$_{300}$ & AIT*$_{1000}$ & BIT*$_{100}$ & BIT*$_{300}$ & BIT*$_{1000}$ & FMT* & Informed RRT* & RRT-Connect & RRT* \\
     \hline\hline
     BugTrap  & 126.9 & 125.0 & 123.4 & 127.1 & 125.4 & 124.6 & 127.6 & 124.9 & 124.8 & 126.7 & 152.4 & 179.4 & 150.6 \\
     Maze     & 146.6 & 145.3 & 145.2 & 146.0 & 145.6 & 145.0 & 146.5 & 145.3 & 145.0 & 147.2 & 160.1 & 206.8 & 167.5 \\
     Easy(3D) & 230.4 & 229.9 & 228.5 & 230.2 & 228.9 & 229.0 & 229.5 & 229.5 & 229.0 & 233.2 & 247.7 & 454.5 & 249.2 \\
     \hline
    \end{tabular}
    }
    \vspace{-0.2cm}
\end{table*}

\begin{figure}[t]
    \centering
    \begin{minipage}{0.31\textwidth}
        \captionsetup[subfloat]{labelformat=empty}
        \subfloat[]{\includegraphics[width=0.98\linewidth]{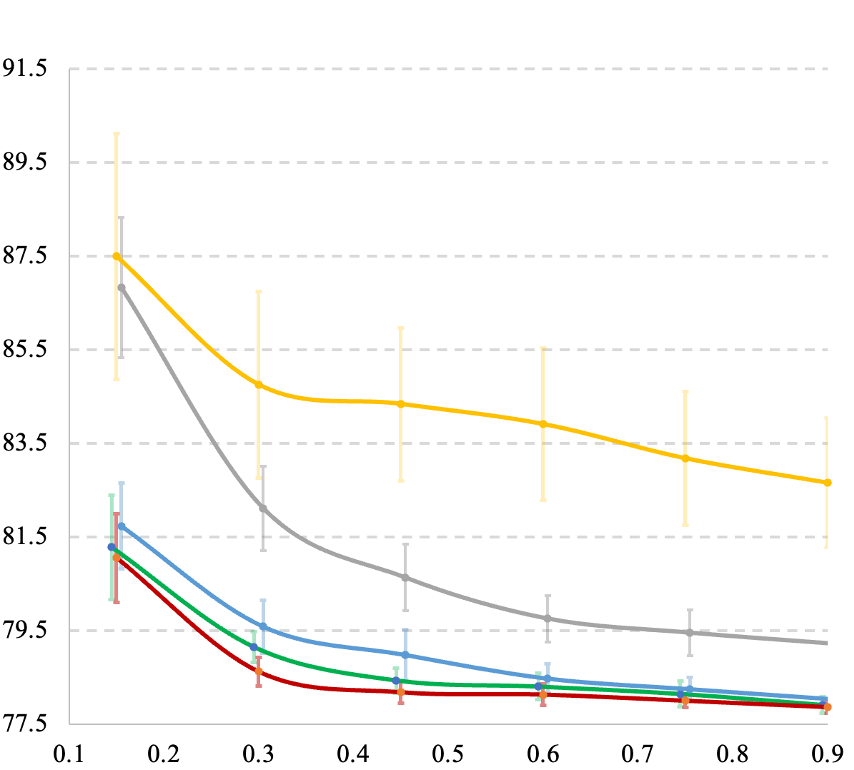}}%
    \end{minipage}
    \begin{minipage}{0.09\textwidth}
        \captionsetup[subfloat]{labelformat=empty}
        \subfloat[]{\includegraphics[width=0.98\linewidth]{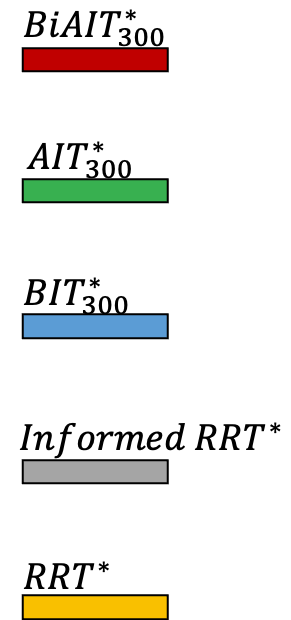}}%
    \end{minipage}
    \hfill
    \caption{Convergence rates of BiAIT*, AIT*, BIT*, Informed RRT*, and RRT*. 
             The figure shows the median values of the solution costs at different time points.
             $batchSize$ of BiAIT*, AIT*, and BIT* is set to $300$.
             The vertical axis denotes the solution quality, and the horizontal axis denotes the computational time.
             }
    \label{OMPLSimResult_Progress}
    \vspace{-0.2cm}
\end{figure}

\begin{figure}[t]
    \centering
    \includegraphics[width=0.36\textwidth]{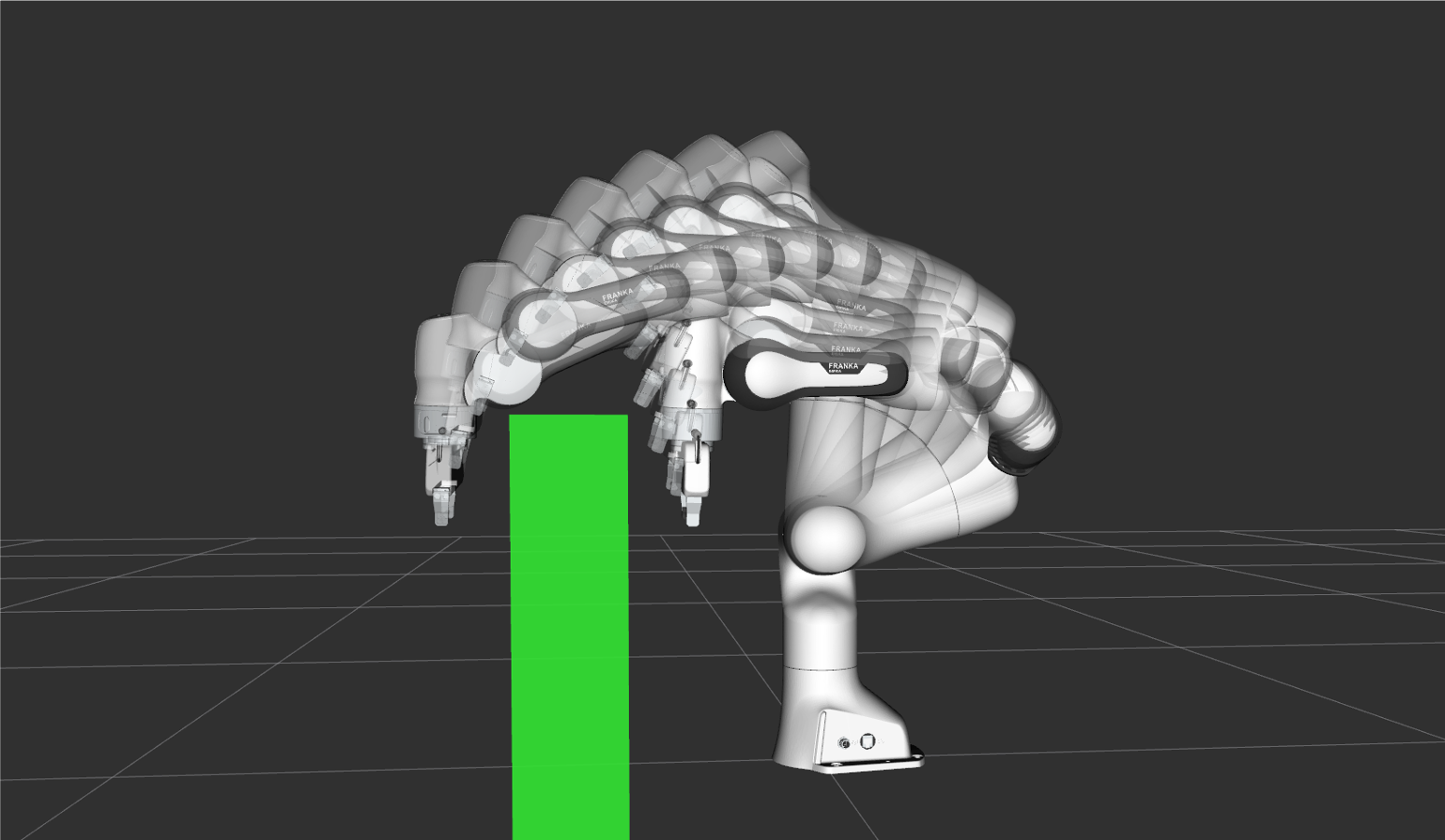}
    \caption{This figure shows the solution path of BiAIT* with a planning time of 0.3 seconds. The green block represents an obstacle. The simulation was conducted on a computer with an Intel Core i9-9900K processor.}
    \label{frankaEmika}
    \vspace{-0.1cm}
\end{figure}

In the simulations, we focus on the speed of reporting the initial solution and the quality of the initial solution.
$batchSize$ is a crucial hyper-parameter for BiAIT*, AIT*, and BIT*. 
Therefore, we set $batchSize$ to different values for comparison; $batchSize$ of BiAIT*, AIT*, and BIT* is set to $100$, $300$, and $1000$ in all scenarios.

Fig. \ref{OMPLSimResult} uses the box plot to show the computational time of finding the initial solution from $100$ independent runs. 
The simulations in $SE(2)$ and $SE(3)$ environments show that BiAIT* outperforms other methods in the speed of finding the initial solution, except slightly worse than the RRT-Connect in the `BugTrap' environment.
$batchSize$ can influence the result, and the optimal $batchSize$ varies in different scenarios. 
In addition to the speed of finding the initial feasible solution, the initial solution quality of BiAIT* is also satisfactory. 
We evaluate the initial solution quality of each planner, and the results are shown in Table \ref{InitialSolutionQualityTable}.
The initial solution quality of BiAIT* stays at the same level as AIT*, BIT*, and FMT*, and is much better than Informed RRT*, RRT-Connect, and RRT*.

\subsection{Comparison of the Convergence Rate}

BiAIT* is an asymptotically optimal planner. 
The convergence rate is an important metric to evaluate its performance.
The simulation environment is a $SE(2)$ space with multiple randomly arranged polygons inside.
BiAIT* is compared with several asymptotical optimal planners.

The simulation result in Fig. \ref{OMPLSimResult_Progress} shows the median values from $100$ independent runs.
The simulation shows that BiAIT* has a faster convergence rate than the other methods. 
While the initial solution quality of BiAIT* is comparable to that of AIT* and BIT*, it can achieve better solution quality given sufficient computational time.

\subsection{Simulation on A 7-DOF Robotic Arm}

Franka Emika robot system is a 7-DOF robotic arm and is widely used in robotic research.
The planning problem is set as planning a path in the joint space ($\mathcal{X} \subset \Re^{7}$) for the Franka Emika while avoiding the obstacle in a limited amount of time.
In this simulation, we compared BiAIT*, AIT*, and BIT*, with their $batchSize$ parameters set to $300$.
We run each planner $50$ times to solve the problem and compare their success rates and solution quality.

Fig. \ref{frankaEmika} shows the path planned by BiAIT*. 
With a planning time of $0.5$ seconds, BiAIT*, AIT*, and BIT* have a $100\%$ success rate in solving the planning problem. 
BiAIT* has a median solution cost of $6.03$, while the median solution costs of AIT* and BIT* are $5.98$ and $6.51$, respectively.
If planners are required to solve the problem in $0.1$ seconds, the success rates of BiAIT*, AIT*, and BIT* are $94\%$, $89\%$, and $71\%$, respectively. 
Their median solution costs are $7.23$, $7.20$, and $8.52$, respectively.
The simulation demonstrates that BiAIT* has a higher success rate than AIT* and BIT* in real-time path planning for the robotic arm. 
Additionally, the solution quality of BiAIT* remains at the same level as that of AIT*.

\section{Experiments}

\begin{figure*}[t]
    \centering
    \begin{minipage}{0.32\textwidth}
        \captionsetup[subfloat]{labelformat=empty}
        \subfloat[{\footnotesize Experiment 1}]{\includegraphics[width=0.98\linewidth]{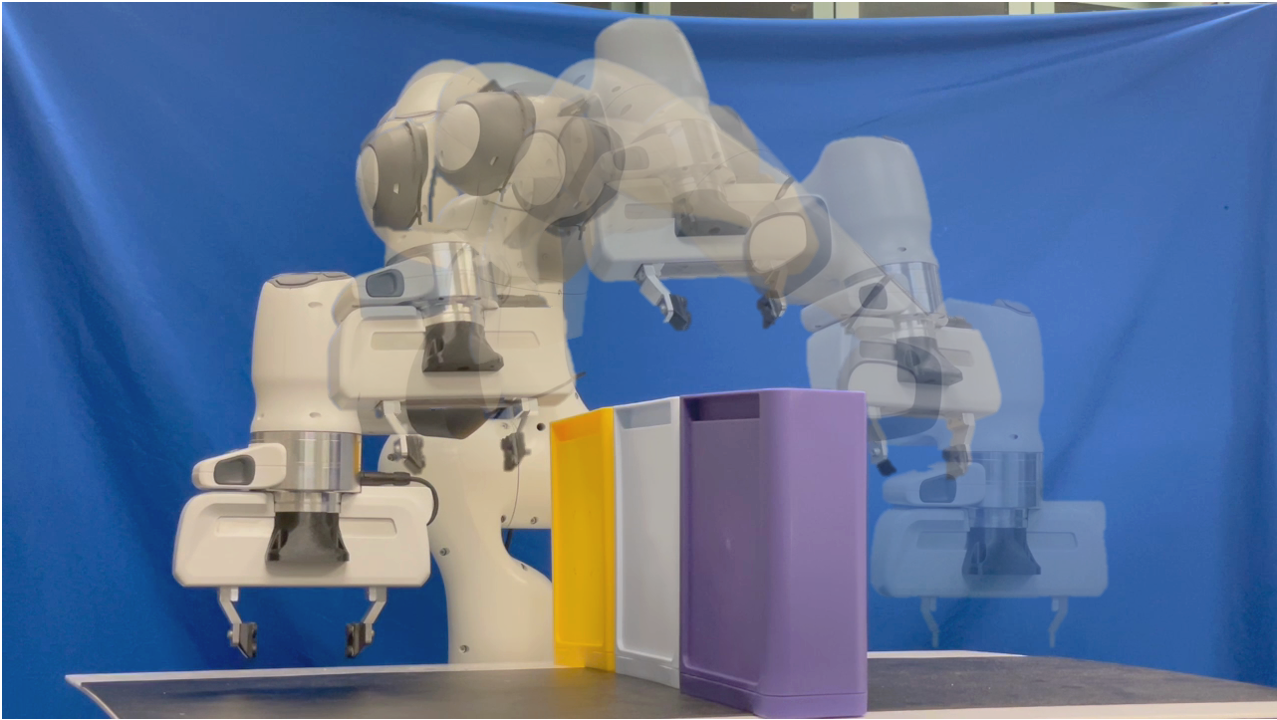}}%
        \label{Exp_1}
    \end{minipage}
    \hfill
    \begin{minipage}{0.32\textwidth}
        \captionsetup[subfloat]{labelformat=empty}
        \subfloat[{\footnotesize Experiment 2}]{\includegraphics[width=0.98\linewidth]{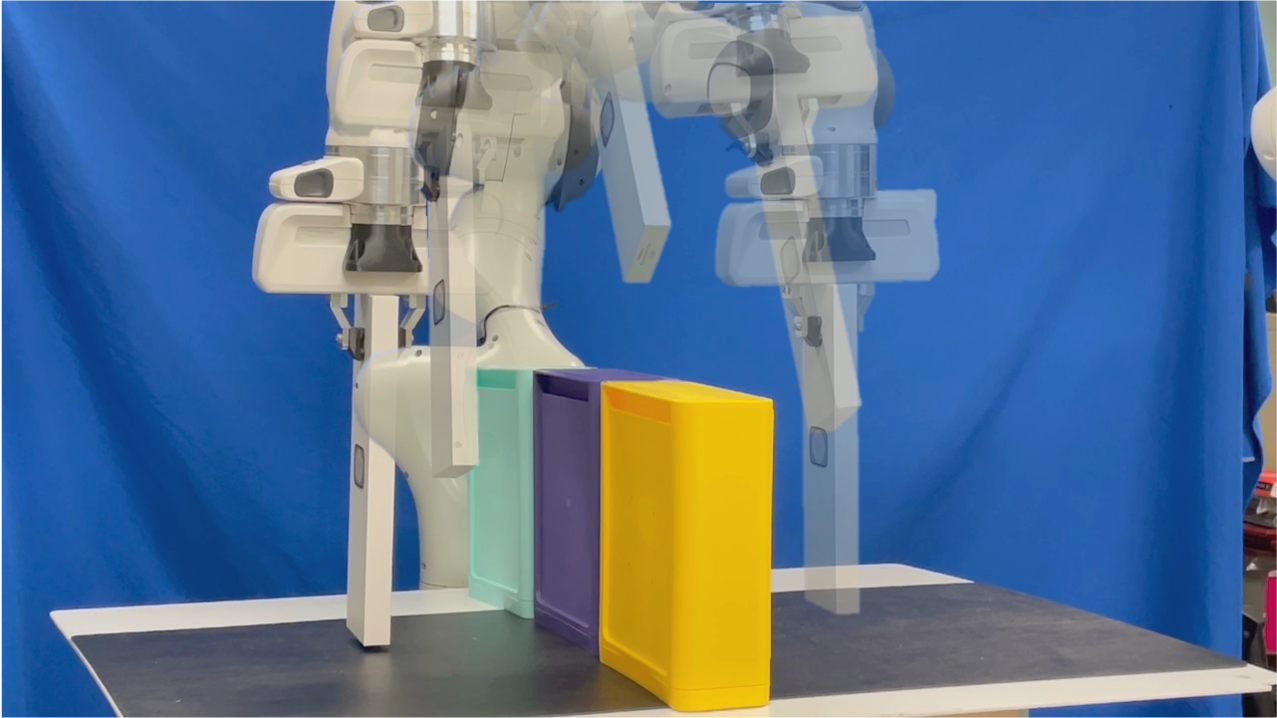}}%
        \label{Exp_2}
    \end{minipage}
    \hfill
    \begin{minipage}{0.32\textwidth}
        \captionsetup[subfloat]{labelformat=empty}
        \subfloat[{\footnotesize Experiment 3}]{\includegraphics[width=0.98\linewidth]{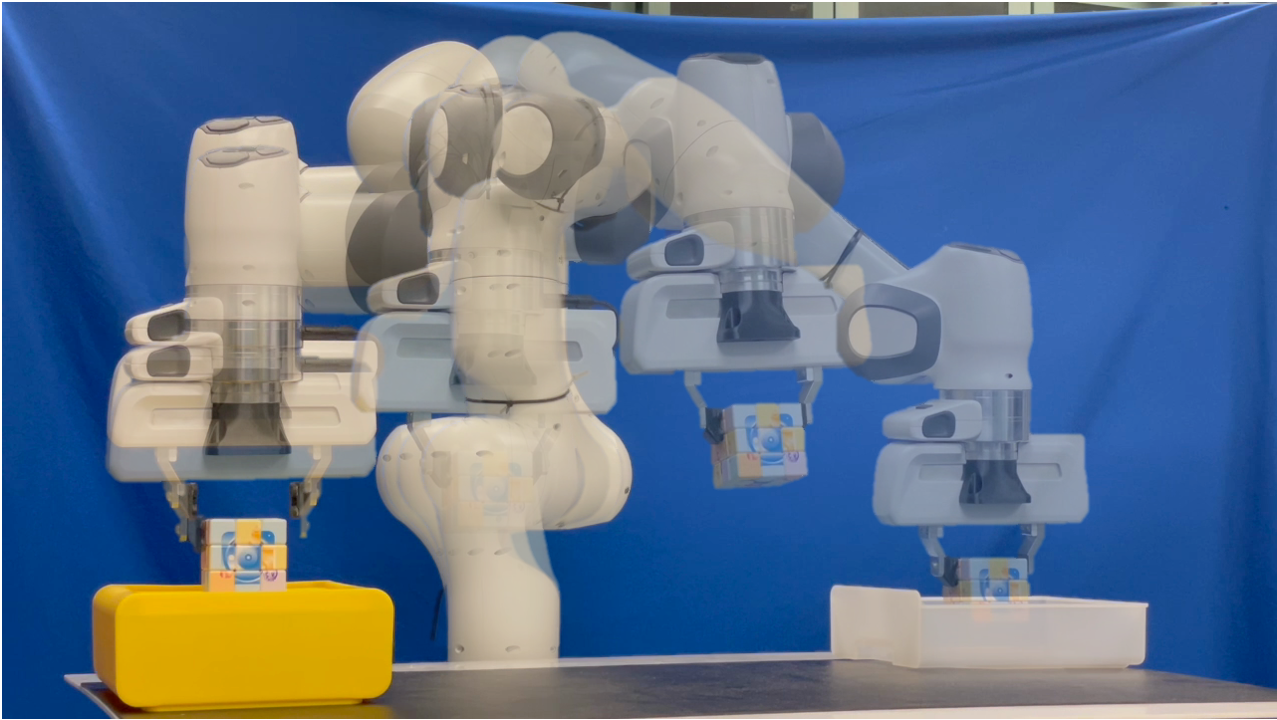}}%
        \label{Exp_3}
    \end{minipage}
    \caption{The experiments consist of three different tasks. The first task involves moving the robotic arm to the goal while avoiding collisions (Experiment 1). The second task involves moving the robotic arm to the goal with an object in hand while avoiding collisions (Experiment 2). The third task is solving a simple manipulation problem (Experiment 3).} 
    \label{ExperimentImg}
    \vspace{-0.2cm}
\end{figure*}

To evaluate the performance of BiAIT* on real-world high-dimensional planning problems, we conduct experiments on Franka Emika. 
The planning problem involved finding a path in joint space ($\mathcal{X} \subset \Re^{7}$). 
We assume that the robot has prior knowledge of the environment since environment perception was not a concern in the experiments. 
We compare the performance of BiAIT* with that of AIT* and BIT*. 
The batch size for all three methods was set to $500$. 
The planned paths are shown in Figure \ref{ExperimentImg}.

In the first experiment, as shown in Fig. \ref{ExperimentImg}, BiAIT*, AIT*, and BIT* are able to find feasible solutions in all $20$ runs, with median solution costs of $6.63$, $6.52$, and $7.68$, respectively, when given $0.5$ seconds of computation time.
In the second experiment, also shown in Fig. \ref{ExperimentImg}, given $0.5$ seconds of computation time, BiAIT* has a $100\%$ success rate in $20$ runs, with a median solution cost of $9.27$, while AIT* and BIT* have success rates of $90\%$ and $75\%$, respectively, with median solution costs of $9.69$ and $10.30$.
In the third experiment, the planning problem consisted of several checkpoints, and planners are tasked with planning the path between two consecutive checkpoints. 
The robotic arm executes and moves along the initial feasible solution. 
The median values of cumulative computation time for the entire path in the third experiment are $0.95s$ (BiAIT*), $1.26s$ (AIT*), and $2.09s$ (BIT*). 
The solution qualities are almost the same for all three planners in the third experiment.
These three experiments show that BiAIT* generally outperforms AIT*, and BIT* in terms of success rate and solution cost.

\section{Analysis}

In this section, firstly, we provide the time complexity analysis.
Then, we prove that $\mathcal{T}_{FL}$ and $\mathcal{T}_{RL}$ in BiAIT* meet in the middle, which is necessary for the resolution-optimal heuristic.
In addition, we explain the reason that BiAIT* consumes less computational cost than AIT* \cite{strub2021ait} with an example.
And we analysis the planning problems that are bad for adaptively informed methods.
Finally, we prove BiAIT* is asymptotically optimal.

\subsection{Complexity Analysis}

The time complexity of BiAIT* comes from three aspects: $Time_{sample}$, $Time_{forward}$, and $Time_{lazySearch}$, which stand for the time complexity of sampling (Algorithm \ref{mainFunction}, Line \ref{pruneLine} to Line \ref{initBatchFunction}), forward search (Algorithm \ref{mainFunction}, Line \ref{mainFunction_ForwardSearch}), and lazy search (Algorithm \ref{mainFunction}, Line \ref{lazySearchFunction}), respectively. 
Sampling is a simple operation. With the sample set of $N$ samples, the time complexity of taking samples is $O(N)$. 
Since BiAIT* uses Geometric Near-neighbor Access Tree (GNAT) \cite{https://doi.org/10.48550/arxiv.1605.05944} to store the sampled points, in the worst case, the time complexity of building a GNAT is $O(N)$. 
However, BiAIT* only builds the GNAT when the newest batch of samples is added, which is rare in the whole process. 
The time complexity of sampling is $Time_{sample} = O(N)$. 
Collision checking dominates the running time of the forward search, $Time_{forward} = T_{collisionChecking} = O(N * log(N))$. 
The aim of BiAIT* is not to simplify the collision checking process but to reduce the chance of calling $forwardSearch()$ (Algorithm \ref{mainFunction}, Line \ref{mainFunction_ForwardSearch}). 
The time complexity of querying a point in GNAT is $Time_{GNAT-Query} = O(log(N))$. Therefore, the time complexity of lazy search $Time_{lazySearch} = Time_{updateState} + log(N)*Time_{updateState}$, where $Time_{updateState} = O(N)$ denotes the time complexity of $updateState()$ (Algorithm \ref{updateStateAlgorithm}). 
Therefore, $Time_{lazySearch} = O(N) + O(N*log(N)) = O(N*log(N))$. 
The time complexity of BiAIT* is $Time_{sample} + Time_{forward} + Time_{lazySearch} = O(N*log(N))$ for $N$ samples, which is equivalent to the time complexity of RRT*. The main contribution of BiAIT* is reducing computational complexity at the constant-term level.

\subsection{Meet-in-the-Middle}     \label{MM_Proof}

We prove the lazy-forward and lazy-reverse searches of BiAIT* can find the resolution-optimal heuristic current knowledge, where
the knowledge includes the valid sample set and the black and white lists, providing a topological abstraction of the environment.
$\hat{c}^*_{\mathcal{X}_s}(x_s, x_g)$ denotes the resolution-optimal heuristic path cost between the start and the goal, where the resolution-optimal heuristic path is acquired from $\mathcal{T}_{FL}$ and $\mathcal{T}_{RL}$ and denoted by $\hat{\pi}^*_{\mathcal{X}_s}(x_s, x_g)$.
In Table. \ref{lexicographicalOrderKey}, $x.\hat{h}_{E\text{-}x_s}$ and $x.\hat{h}_{E\text{-}x_g}$ are the straight-line Euclidean distance to $x_s$ and $x_g$, respectively, which is the lower bound of the heuristic, $x.\hat{h}_{E\text{-}x_s} \leq \min(x.\hat{h}_{g\text{-}F},\ x.\hat{h}_{rhs\text{-}F})$ and $x.\hat{h}_{E\text{-}x_g} \leq \min(x.\hat{h}_{g\text{-}R},\ x.\hat{h}_{rhs\text{-}R})$. 
Therefore, the straight-line Euclidean distance to $x_s$ and $x_g$ can be viewed as the admissible heuristic for the lazy-forward and lazy-reverse searches:
\begin{equation}
    \begin{split}
        &x.\hat{h}_{E\text{-}x_g} + \min(x.\hat{h}_{g\text{-}F},\ x.\hat{h}_{rhs\text{-}F}) \leq \hat{c}^*(x_s, x_g) \\
        &x.\hat{h}_{E\text{-}x_s} + \min(x.\hat{h}_{g\text{-}R},\ x.\hat{h}_{rhs\text{-}R}) \leq \hat{c}^*(x_s, x_g).
    \end{split}
\end{equation}

To enable $\mathcal{T}_{FL}$ and $\mathcal{T}_{RL}$ meet in the middle of $\hat{\pi}^*_{\mathcal{X}_s}(x_s, x_g)$, BiAIT* limits the lazy-forward and lazy-reverse searches explore the vertex whose $\min(x.\hat{h}_{g\text{-}F},\ x.\hat{h}_{rhs\text{-}F}) \leq \hat{c}^*_{\mathcal{X}_s}(x_s, x_g)/2$ and $\min(x.\hat{h}_{g\text{-}R},\ x.\hat{h}_{rhs\text{-}R}) \leq \hat{c}^*_{\mathcal{X}_s}(x_s, x_g)/2$, respectively.
Therefore, all vertices explored by lazy-forward and lazy-reverse searches have $2*\min(x.\hat{h}_{g\text{-}F},\ x.\hat{h}_{rhs\text{-}F}) \leq \hat{c}^*_{\mathcal{X}_s}(x_s, x_g)$ and $2*\min(x.\hat{h}_{g\text{-}R},\ x.\hat{h}_{rhs\text{-}R}) \leq \hat{c}^*_{\mathcal{X}_s}(x_s, x_g)$, respectively.
Since the frontiers of lazy-forward and lazy-reverse searches are sorted according to the keys in Table. \ref{lexicographicalOrderKey}, BiAIT* does not explore any vertex whose $g$-value or $rhs$-value is greater than $\hat{c}^*_{\mathcal{X}_s}(x_s, x_g)/2$.
In addition, the meet vertex of $\mathcal{T}_{FL}$ and $\mathcal{T}_{RL}$ in current batch of samples satisfies $\min(x.\hat{h}_{g\text{-}F},\ x.\hat{h}_{rhs\text{-}F}) + \min(x.\hat{h}_{g\text{-}R},\ x.\hat{h}_{rhs\text{-}R}) = \hat{C}^*_{\mathcal{X}_s}(x_s, x_g)$.
In summary, meeting in the middle enables BiAIT* to generate the resolution-optimal heuristic under current knowledge.

To enable $\mathcal{T}_{F}$ and $\mathcal{T}_{R}$ to meet in the middle, BiAIT* constraints $\mathcal{T}_{F}$ and $\mathcal{T}_{R}$ to the $\mathcal{T}_{FL}$ and $\mathcal{T}_{RL}$ explored region, respectively.


\begin{figure}[t]
    \centering
    \begin{minipage}{0.23\textwidth}
        \begin{subfloat}[]{\includegraphics[width=0.98\linewidth]{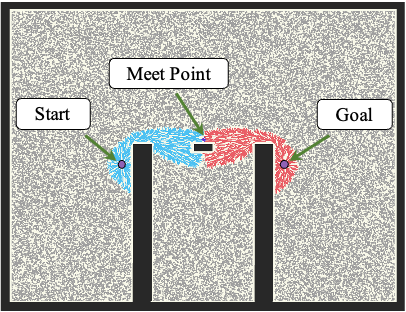}}%
        \end{subfloat}
        \label{Fig_MM}
    \end{minipage}
    \hfill
    \begin{minipage}{0.23\textwidth}
        \begin{subfloat}[]{\includegraphics[width=0.98\linewidth]{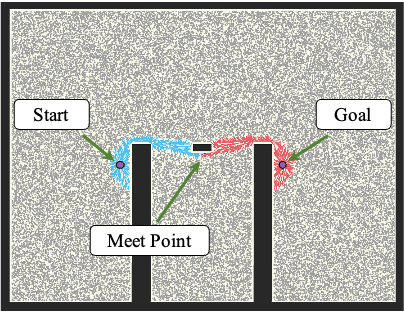}}%
        \end{subfloat}
        \label{Fig_No_MM}
    \end{minipage}
    \hfill
    \begin{minipage}{0.23\textwidth}
        \begin{subfloat}[]{\includegraphics[width=0.98\linewidth]{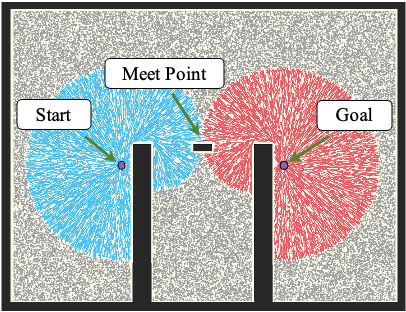}}%
        \end{subfloat}
        \label{Fig_No_H}
    \end{minipage}
    \hfill
    \begin{minipage}{0.23\textwidth}
        \begin{subfloat}[]{\includegraphics[width=0.98\linewidth]{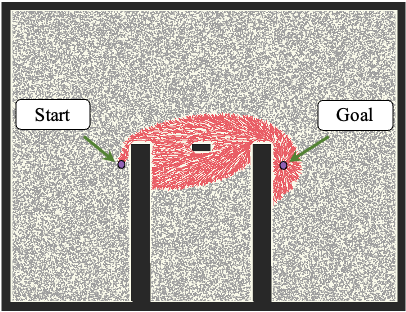}}%
        \end{subfloat}
        \label{Fig_AIT}
    \end{minipage}

    \caption{
        The figure shows the space explored by the various lazy search strategies in BiAIT* (Fig. \ref{CompareLazyExploreSpace}a, \ref{CompareLazyExploreSpace}b, and \ref{CompareLazyExploreSpace}c) as well as the lazy-reverse search in AIT* (\ref{CompareLazyExploreSpace}d). 
        Obstacles, samples, the start point, and the goal point are denoted by black blocks, grey points, and violet points, respectively.
        The lazy-reverse tree is shown in red, while the lazy-forward tree is shown in blue.
        }
    \label{CompareLazyExploreSpace}
\end{figure}

\subsection{Necessity of Meet-in-the-Middle in Lazy Searches}


We illustrate why BiAIT* chooses to enable $\mathcal{T}_{FL}$ and $\mathcal{T}_{RL}$ to meet in the middle with the example shown in Fig. \ref{CompareLazyExploreSpace}. 
In Fig. \ref{CompareLazyExploreSpace}a, two lazy trees ($\mathcal{T}_{FL}$ and $\mathcal{T}_{RL}$) of BiAIT* simultaneously propagate wavefronts through the $\mathcal{X}_{free}$ and meet in the middle of the resolution-optimal heuristic path under current knowledge. 
Fig. \ref{CompareLazyExploreSpace}b shows the region explored by the bidirectional lazy search strategy without MM, where $\mathcal{T}_{FL}$ and $\mathcal{T}_{RL}$ meet under the central obstacle. 
That is, the heuristic information is not resolution-optimal without the MM strategy, since $\mathcal{T}_{FL}$ and $\mathcal{T}_{RL}$ tend to explore along different paths and fail to meet in the middle of the resolution-optimal heuristic path. 
Fig. \ref{CompareLazyExploreSpace}c shows the region explored by the brute-force bidirectional lazy search strategy, where the meet point of $\mathcal{T}_{FL}$ and $\mathcal{T}_{RL}$ is located in the middle of the resolution-optimal path with plenty of irrelevant space explored by $\mathcal{T}_{FL}$ and $\mathcal{T}_{RL}$.
In summary, Fig. \ref{CompareLazyExploreSpace}a-c show that BiAIT* with the MM strategy outperforms the brute-force exploring method and the heuristic searching method without MM. 
Specifically, the way we define $x.key^{BiAIT*}_{\mathcal{Q}_{FL}}$ and $x.key^{BiAIT*}_{\mathcal{Q}_{RL}}$ guarantees the efficiency and resolution-optimality of the lazy searches.


\subsection{Advantages over AIT*}

Figure \ref{CompareLazyExploreSpace}d shows the space explored by the lazy-reverse tree of AIT*. 
As depicted in Figures \ref{CompareLazyExploreSpace}a and \ref{CompareLazyExploreSpace}d, the lazy search of BiAIT* explores fewer vertices than AIT* to obtain the initial heuristic. 
However, we note that there are certain scenarios where the lazy search of AIT* can explore fewer vertices than BiAIT*.
Another advantage of BiAIT* over AIT* is the $updateLazySearch()$ function (Algorithm \ref{AITAlgorithm}, Line \ref{updateLazySearchingLineInAIT} and Algorithm \ref{validSearchAlgorithm}, Line \ref{updateLazySearchLine}). 
$updateLazySearch()$ in AIT* and BiAIT* updates and invalidates the affected lazy branch when a collision occurs. 
In AIT*, the subtree $\mathcal{T}_{inv}$ rooted at the child vertex of the collided edge ($x_c \in \lbrace x_p, x_c \rbrace$ and $\lbrace x_p, x_c \rbrace \cap \mathcal{X}^{obs} \neq \emptyset$) needs to be invalidated. 
$\mathcal{T}_{inv}$ typically covers the entire extended space from $x_c$ to the goal region $\mathcal{X}^g$ in AIT*. 
In contrast, in BiAIT*, $\mathcal{T}_{inv}$ covers the space from $x_c$ to its descendants $x_d \in \mathcal{S}^{lazyMeet}$, which results in a smaller $\mathcal{T}_{inv}$ than that of AIT*.
In addition to invalidating $\mathcal{T}_{inv}$, BiAIT* needs to propagate the change to the start and the goal (Algorithm \ref{updateLazySearchAlgorithm}, Lines \ref{lineUpdatePredecessorLineToStart} and \ref{lineUpdatePredecessorToGoal}).
When BiAIT* propagates the change in $\mathcal{T}_{FL}$ and $\mathcal{T}_{RL}$, as shown in Fig. \ref{diffInInvalidation}, the structure of $\mathcal{T}_{FL}$ and $\mathcal{T}_{RL}$ is maintained, resulting in less computational consumption when a collision occurs.

\subsection{The Thickness of Obstacles}
The thickness of obstacle walls is a key factor that affects the performance of adaptively informed methods. 
In such cases, edges in $\mathcal{T}_{FL}$ and $\mathcal{T}_{RL}$ may have a higher probability of crossing the obstacle walls, which results in many redundant collision checks in forward and reverse searches. 
We notice that this situation can occur, which reduces calculation speed and solution quality. 
However, in this limiting case, BiAIT* can reduce the impact of this problem from three aspects.
Firstly, in the planning procedure, BiAIT* iteratively adds new samples to enhance the abstraction level of the environment. 
Specifically, there are no valid samples in the obstacle area despite the thinness of the obstacle, therefore, the environment can be correctly abstracted by sufficient samples.
Secondly, BiAIT* uses black and white lists to enable $\mathcal{T}_{FL}$, $\mathcal{T}_{RL}$, $\mathcal{T}_{F}$, and $\mathcal{T}_{R}$ to remember which vertex is unreachable, thus avoiding repeated collision checking. 
That is, if a lazy connection is found to collide with the obstacles during the forward or reverse searches, this lazy connection will not be considered in the following search process.
Finally, BiAIT* treats vertices within a radius ($r$) as neighbors of the vertex ($x$). The calculation of $r$ is shown in (\ref{Equation_NearNeighbors}).
This means that when the number of samples is large enough, the situation where lazy trees pass through the wall can be avoided since there are no samples in obstacles and the radius ($r$) is relatively small.

In addition, we also proposed to increase the number of samples exponentially in Section VIII.A, which can effectively solve this type of problem.

\subsection{Asymptotical Optimality}
The increasingly dense random geometric graph in BiAIT* includes all edges in PRM* for any set of samples \cite{penrose2003random}. 
As a result, the random geometric graph provides the asymptotically optimal solution as the number of samples approaches infinity, since PRM* is an almost-surely asymptotically optimal planner.
LPA* is also a resolution-optimal method, and we have proven that the bidirectional version of LPA*'s incrementally search strategy can result in the resolution-optimal heuristic path under current knowledge in Section \ref{MM_Proof}. 
Specifically, the heuristic path is resolution-optimal since the lazy-forward and lazy-reverse trees meet in the middle and they are guided by the Euclidean-metric-based front-to-end heuristic.
The lazy-forward and lazy-reverse searches of BiAIT* provide the admissible heuristic to the forward and reverse searches since the bidirectional lazy search strategy is a resolution-optimal method and including the collision checking process does not reduce the heuristic path cost in the lazy-forward and lazy-reverse searches. 
The forward and reverse searches of BiAIT* are resolution-optimal with admissible heuristic, as the MM \cite{holte2017mm} algorithm is a resolution-optimal algorithm. 
Therefore, BiAIT* is an almost-surely asymptotically optimal planner.

\section{Modifications}
We briefly discuss two modifications in this section, which are applicable to both AIT* and BiAIT*. These two modifications can help to improve the planner to achieve less computational consumption.

\begin{figure}[t]
    \centering
    \hfill
    \begin{minipage}{0.32\textwidth}\centering
        \captionsetup[subfloat]{labelformat=empty}
        \subfloat[]{\includegraphics[width=0.98\linewidth]{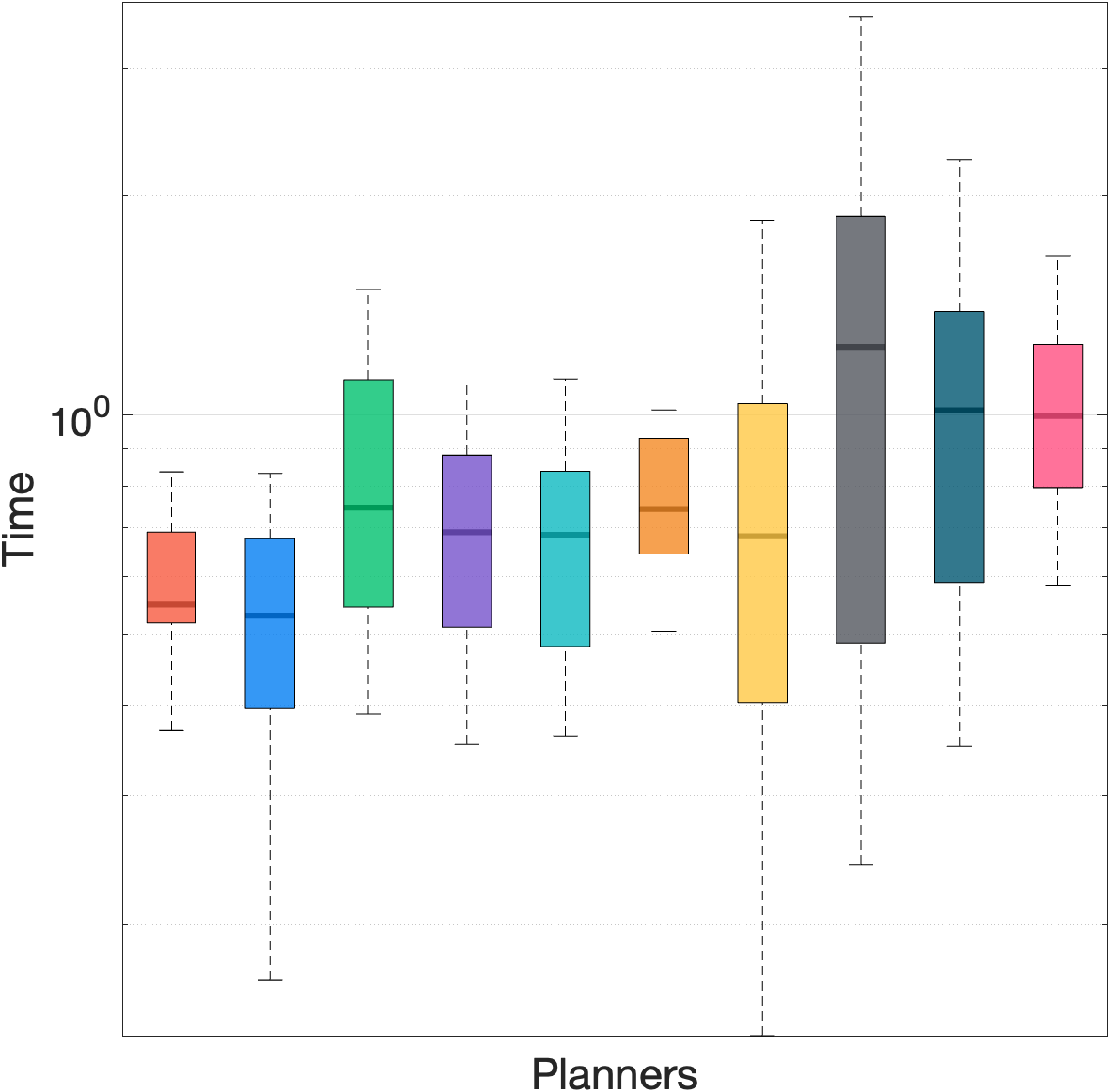}}%
    \end{minipage}
    \hfill
    \begin{minipage}{0.115\textwidth}\centering
        \captionsetup[subfloat]{labelformat=empty}
        \subfloat[]{\includegraphics[width=0.98\linewidth]{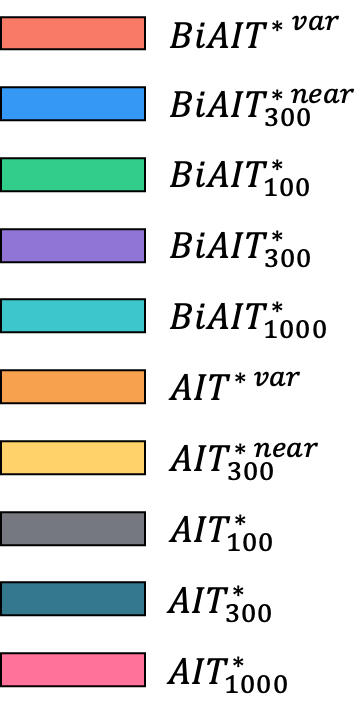}}%
    \end{minipage}
    \hfill
    \caption{Comparison of  the two modifications against BiAIT* and AIT* ($100$ runs). The path planning problem is finding a path within $5\%$ tolerance of the optimal path in OMPL 'BugTrap' $SE(2)$ environment. $BiAIT^{*var}$ and $AIT^{*var}$ use a variational batch size with $init = 10$ and $\alpha=1.5$. The $BiAIT^{*near}_{300}$ and $AIT^{*near}_{300}$ samples near the current optimal path with $batchSize = 300$ and $p_{near} = 0.5$.}
    \label{variationalBatchSize}
    \vspace{-0.2cm}
\end{figure}

\subsection{Variational Batch Size}

Both AIT* and BiAIT* use a constant batch size to update the heuristic.
However, we find that the sampling and the following lazy search provide almost no better heuristic when the batch number approaches $\infty$.
Let $n$ denote the batch number. 
The topological abstraction ablity of $n$ batches of samples almost equals to $n+1$ batches of samples ($n \to \infty$).
In addition, it is not easy to determine an optimal $batchSize$ for different planning problems.
AIT* and BiAIT* use a constant $batchSize$ as the planning problem changes.
Therefore, we propose to let $batchSize$ increase exponentially instead of using a constant $batchSize$.
The $n$-th batch size is $init * (1+\alpha)^n$, where the $init$ is the initial batch size, which is a relatively small number.
We varified this idea and tested its performance in a simulation environment. 
The simulation results are shown in Fig. \ref{variationalBatchSize}.

\subsection{Nonuniform Sampling}

The informed sampling method acquires a compact sampling region while almost does not increase the computational burden.
But the informed sampling region is not compact enough, especially when the theoretical minimum cost is small and the current optimal solution cost is large.
Sampling near the current optimal path can help find the optimal homotopy solution faster and result in a more compact sampling region.
We propose to sample near the current optimal path with a probability $p_{near} \in [0, 1)$, and sample uniformly in the informed region with probability $1-p_{near}$.
Since we keep sampling in the informed region with probability $1-p_{near} > 0$, the probabilistic completeness and global asymptotical optimality are preserved.
However, we have to admit that the nonuniform sampling will reduce the chance of finding the non-homotopy optimal solution.
The simulation results are shown in Fig. \ref{variationalBatchSize}.
The results also show BiAIT* performs better in the convergence speed than AIT*.


\section{Conclusions}
We develop BiAIT*, which extends the adaptively heuristic searching from the asymmetric bidirectional method to the symmetrical bidirectional method.
BiAIT* constructs the lazy-forward and lazy-reverse trees to acquire the problem-specific adaptively heuristic information.
The lazy trees share the heuristic information when they meet.
If either the forward search or the reverse search may improve current $\mathcal{T}_{F}$, $\mathcal{T}_{R}$, or $c_{cur}$, BiAIT* will attempt to construct $\mathcal{T}_{F}$ or $\mathcal{T}_{R}$.
Forward and reverse searches will inform $\mathcal{T}_{FL}$ or $\mathcal{T}_{RL}$ and update the heuristic when collisions occur.
Simulations in $SE(2)$ and $SE(3)$ spaces demonstrate that BiAIT* can find an initial solution with satisfactory quality in less time.
We also verify BiAIT* with Franka Emika in both simulaiton and real-world experiments.
Additionally, we discuss the computational complexity, MM strategy for lazy searches, and the superiorities of BiAIT* in depth.
We also propose two simple but effective modifications which are applicable to both AIT* and BiAIT*.

\section{Appendix}

In Algorithm \ref{AlgFuncsFromAIT} shows the details of $sample()$, $expand(x)$, and $prune(\mathcal{X}_{s}, \mathcal{T}_{F}, \mathcal{T}_{R})$.
These functions are similar to the corresponding parts in AIT* \cite{strub2021ait}, and we include them here for completeness.
$sample()$ takes a batch of samples in the informed space when $c_{cur}$ is not infinite; otherwise, $sample()$ samples uniformly.
After the feasible solution has been found, $sample()$ first samples from the $n$-dimensional unit ball centered at the origin point.
These samples are then transfered to the $n$-dimensional hyper-spheroid (Algorithm \ref{AlgFuncsFromAIT}, Line \ref{lineRemapToHPS}), where $C$ is a rotation matrix transfering the samples to the world frame.
The function $expand(x)$ (Algorithm \ref{AlgFuncsFromAIT}, Line \ref{funcExpand}) outputs the outgoing edges of $x$ without considering its blacklisted neighbors, which are used in the forward and reverse searches to improve the $\mathcal{T}_{F}$ or $\mathcal{T}_{R}$.
The function $prune(\mathcal{X}_{s}, \mathcal{T}_{F}, \mathcal{T}_{R})$ (Algorithm \ref{AlgFuncsFromAIT}, Line \ref{funcPrune}) removes samples from $\mathcal{X}_{s}$ and branches from $\mathcal{T}_{F}$ and $\mathcal{T}_{R}$ that cannot improve the current solution.

\begin{algorithm}[t]
    \caption{Functions From AIT* \cite{strub2021ait}}
    \label{AlgFuncsFromAIT}
    \SetKwFunction{sample}{$sample$}
    \SetKwFunction{expand}{$expand$}
    \SetKwFunction{prune}{$prune$}
    \SetKwProg{CustomFunc}{Function}{}{}
    \CustomFunc{\sample$()${}}{         \label{funcSample}
        \If{$c_{cur} < \infty$}{
            $r_1 \gets c_{cur}$\;
            $\lbrace r_i \rbrace_{i = 2,\dots,n} \gets \frac{\sqrt{c_{cur}^2 - \lVert x_{g} - x_{s} \rVert^{2}}}{2}$\;
            $\boldmath{L} \gets diag(\lbrace r_1, r_2, \dots, r_n \rbrace)$\;
            $\mathcal{X}_{unitBall} \gets sampleUniteBall()$\;
            $\mathcal{X}_{new} \gets (\boldmath{CL}\mathcal{X}_{unitBall} + \frac{x_{s} + x_{g}}{2})$\; \label{lineRemapToHPS}
        }
        \Else {
            $\mathcal{X}_{new} \gets sampleUniformly()$\;
        }
        \Return $\mathcal{X}_{new}$\;
    }
    \CustomFunc{\expand$(x)${}}{    \label{funcExpand}
        $\mathcal{E}_{out} \gets \emptyset$\;
        \ForAll{$x_n \in x._{neighbors} \backslash x._{black}$}{
            $\mathcal{E}_{out}.insert(\lbrace x, x_n \rbrace)$\;
        } 
        \Return $\mathcal{E}_{out}$\;
    }
    \CustomFunc{\prune$(\mathcal{X}_{s}, \mathcal{T}_{F}, \mathcal{T}_{R})${}}{     \label{funcPrune}
        $\mathcal{X}_{prune} \gets \lbrace x \in \mathcal{X}_{s} | x.\hat{g}_{F} + x.\hat{g}_{R} > c_{cur} \rbrace$\;
        $\mathcal{X}_{s} \gets \mathcal{X}_{s} \backslash \mathcal{X}_{prune}$\;
        \ForAll{$x \in \mathcal{X}_{prune}$}{
            \If{$x.\hat{g}_{F} + x.\hat{g}_{R} > c_{cur}$}{
                $\mathcal{T}_{F}.remove(\lbrace x._{parent}, x\rbrace \cup \lbrace x, x._{children}\rbrace)$\;
                $\mathcal{T}_{R}.remove(\lbrace x._{parent}, x\rbrace \cup \lbrace x, x._{children}\rbrace)$\;
            }
        }
    }       \label{funcPruneEnd}
\end{algorithm}

\bibliographystyle{IEEEtran}
\bibliography{References.bib}

\begin{IEEEbiography}[{\includegraphics[width=1in,height=1.25in,clip,keepaspectratio]{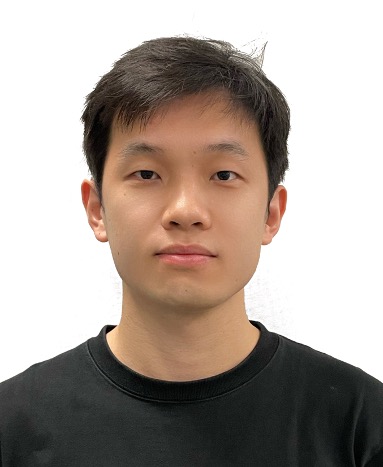}}]{Chenming Li}
    received the B.E. degree in petroleum engineering from the China University of Petroleum, Qingdao, China, in 2017, and the M.Sc. degree in electronic engineering from The Chinese University of Hong Kong, Hong Kong SAR, China, in 2018. He is currently pursuing the Ph.D. degree with the Department of Electronic Engineering, The Chinese University of Hong Kong. His current research interest is robot planning algorithms.
\end{IEEEbiography}

\begin{IEEEbiography}[{\includegraphics[width=1in,height=1.25in,clip,keepaspectratio]{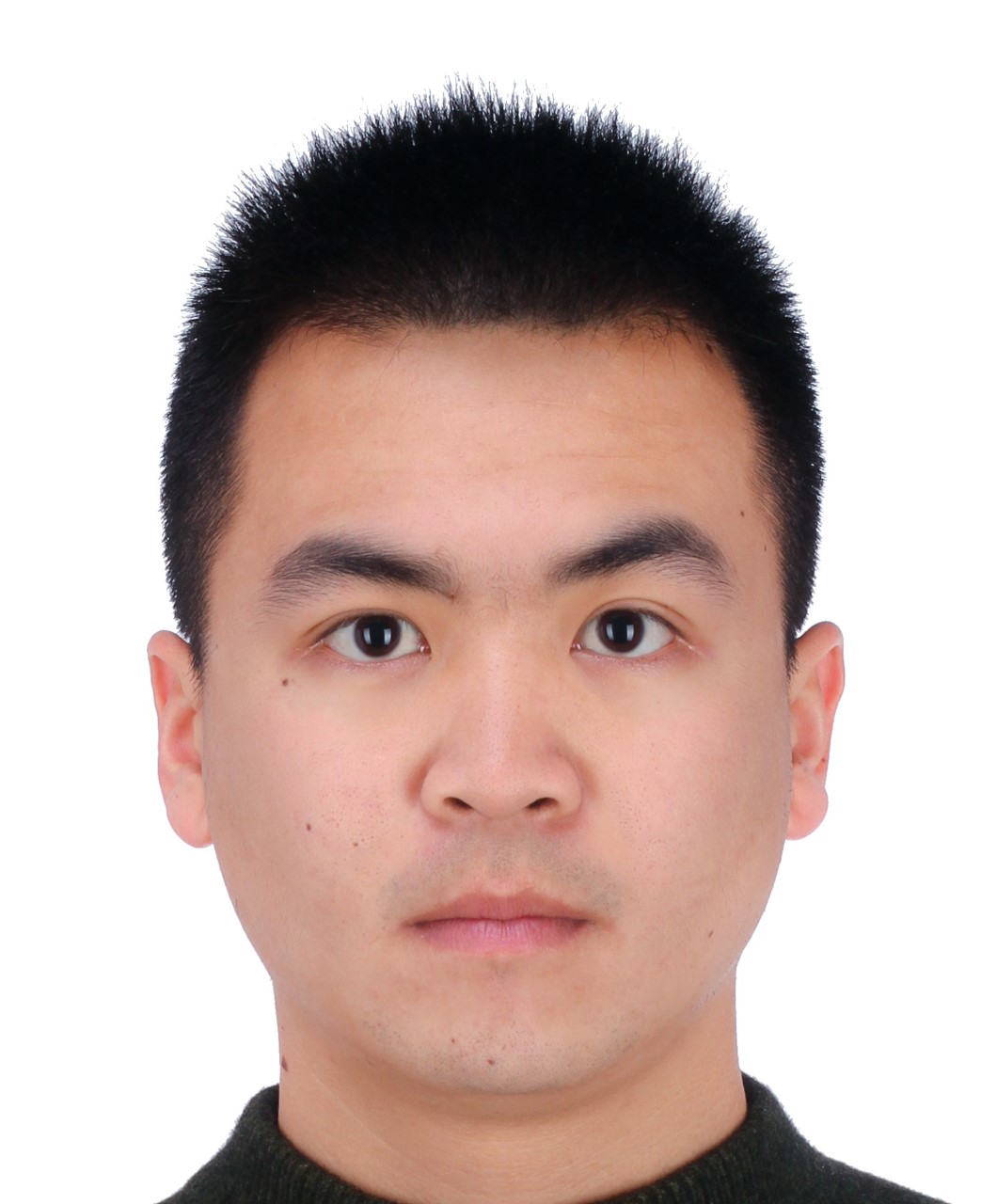}}]{Han Ma}
    received the B.E. degree in measurement, control technology and instrument from the Department of Precision Instrument of Tsinghua University, Beijing, China, in 2019. He is now working towards the Ph.D. degree in the Department of Electronic Engineering of The Chinese University of Hong Kong, Hong Kong SAR, China. His research interests include path planning, deep learning.
\end{IEEEbiography}

\begin{IEEEbiography}[{\includegraphics[width=1in,height=1.25in,clip,keepaspectratio]{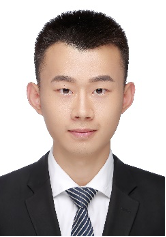}}]{Peng Xu}
    received the B.E. degree in measurement, control technology and instrument from the Department of Precision Instrument of Tsinghua University, Beijing, China, in 2018. He is now working towards the Ph.D. degree in the Department of Electronic Engineering of The Chinese University of Hong Kong, Hong Kong SAR, China. His research interests include Robotics, Deep Learning.
\end{IEEEbiography}

\begin{IEEEbiography}[{\includegraphics[width=1in,height=1.25in,clip,keepaspectratio]{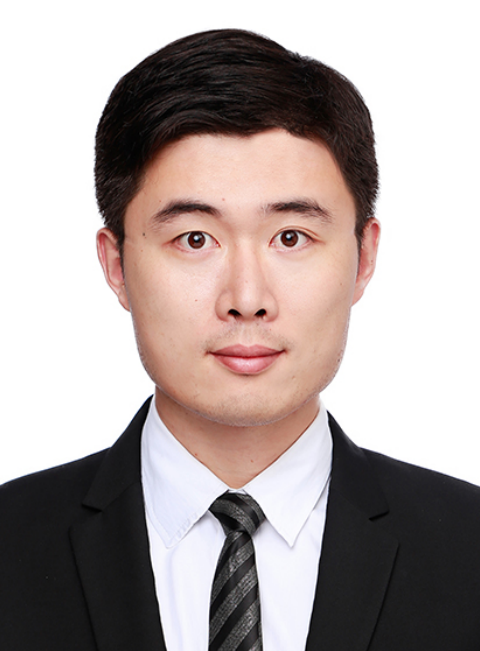}}]{Jiankun Wang}

    (Member, IEEE) received the B.E. degree in automation from Shandong University, Jinan, China, in 2015, and the Ph.D. degree from the Department of Electronic Engineering, The Chinese University of Hong Kong, Hong Kong, in 2019.

    During his Ph.D. degree, he spent six months at Stanford University, Stanford, CA, USA, as a Visiting Student Scholar, supervised by Prof. Oussama Khatib. He is currently a Research Assistant Professor with the Department of Electronic and Electrical Engineering, Southern University of Science and Technology, Shenzhen, China. His current research interests include motion planning and control, human–robot interaction, and machine learning in robotics.
\end{IEEEbiography}

\begin{IEEEbiography}[{\includegraphics[width=1in,height=1.25in,clip,keepaspectratio]{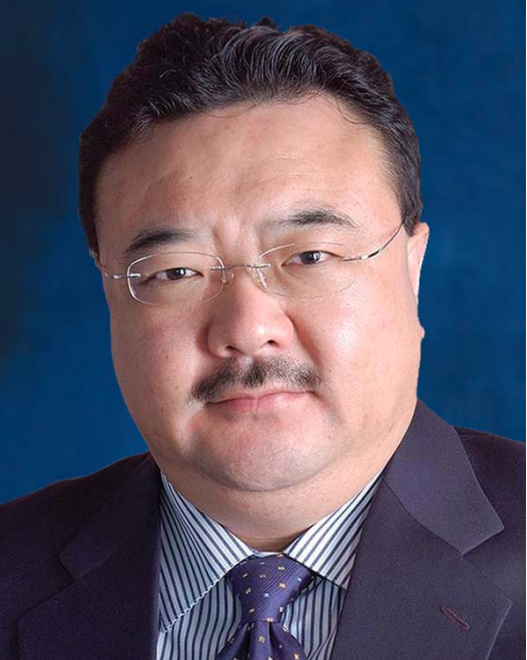}}]{Max Q.-H. Meng}
    (Fellow, IEEE) received his Ph.D. degree in Electrical and Computer Engineering from the University of Victoria, Canada, in 1992. He is currently a Chair Professor and the Head of the Department of Electronic and Electrical Engineering at the Southern University of Science and Technology in Shenzhen, China, on leave from the Department of Electronic Engineering at the Chinese University of Hong Kong. He joined the Chinese University of Hong Kong in 2001 as a Professor and later the Chairman of Department of Electronic Engineering. He was with the Department of Electrical and Computer Engineering at the University of Alberta in Canada, where he served as the Director of the ART (Advanced Robotics and Teleoperation) Lab and held the positions of Assistant Professor (1994), Associate Professor (1998), and Professor (2000), respectively. He is an Honorary Chair Professor at Harbin Institute of Technology and Zhejiang University, and also the Honorary Dean of the School of Control Science and Engineering at Shandong University, in China. 

    His research interests include medical and service robotics, robotics perception and intelligence. He has published more than 750 journal and conference papers and book chapters and led more than 60 funded research projects to completion as Principal Investigator. 

    Prof. Meng has been serving as the Editor-in-Chief and editorial board of a number of international journals, including the Editor-in-Chief of the Elsevier Journal of Biomimetic Intelligence and Robotics, and as the General Chair or Program Chair of many international conferences, including the General Chair of IROS 2005 and ICRA 2021, respectively. He served as an Associate VP for Conferences of the IEEE Robotics and Automation Society (2004-2007), Co-Chair of the Fellow Evaluation Committee and an elected member of the AdCom of IEEE RAS for two terms. He is a recipient of the IEEE Millennium Medal, a Fellow of IEEE, a Fellow of Hong Kong Institution of Engineers, and an Academician of the Canadian Academy of Engineering.
    
\end{IEEEbiography}


\balance

\end{document}